\documentclass{article}

\PassOptionsToPackage{numbers,sort&compress}{natbib}
\usepackage[final]{neurips_2023}

\usepackage[utf8]{inputenc} 
\usepackage[T1]{fontenc}    
\usepackage{hyperref}       
\usepackage{url}            
\usepackage{booktabs}       
\usepackage{amsfonts}       
\usepackage{nicefrac}       
\usepackage{microtype}      
\usepackage{xcolor}         
\usepackage{gensymb}
\usepackage{adjustbox}
\usepackage{wrapfig}
\usepackage{multirow}
\usepackage{rotating}
\usepackage{float}

\title{Balancing Computational Efficiency and Forecast Error in Machine Learning-based Time-Series Forecasting: Insights from Live Experiments on Meteorological Nowcasting}

\author{%
  Elin Törnquist\thanks{\url{https://emergentmethods.ai}} \\
  Emergent Methods\\
  Malmö, Sweden \\
  \texttt{elin@emergentmethods.ai} \\
  \And
  Wagner Costa Santos \\
  Emergent Methods\\
  Sao Paulo, Brazil \\
  \texttt{wagner@emergentmethods.ai} \\
  \And
  Timothy Pogue \\
  Emergent Methods\\
  Colorado, USA \\
  \texttt{tim@emergentmethods.ai} \\
  \And
  Nicholas Wingle \\
  WingleWARE\\
  Michigan, USA \\
  \texttt{nick@wingleware.com} \\
  \And
  Robert A. Caulk \\
  Emergent Methods\\
  Grenoble, France\\
  \texttt{rob@emergentmethods.ai} \\
}

\begin{document}

\maketitle

\begin{abstract}
Machine learning for time-series forecasting remains a key area of research. Despite successful application of many machine learning techniques, relating computational efficiency to forecast error remains an under-explored domain. This paper addresses this topic through a series of real-time experiments to quantify the relationship between computational cost and forecast error using meteorological nowcasting as an example use-case. We employ a variety of popular regression techniques (XGBoost, FC-MLP, Transformer, and LSTM) for multi-horizon, short-term forecasting of three variables (temperature, wind speed, and cloud cover) for multiple locations. During a 5-day live experiment, 4000 data sources were streamed for training and inferencing 144 models per hour. These models were parameterized to explore forecast error for two computational cost minimization methods: a novel auto-adaptive data reduction technique (Variance Horizon) and a performance-based concept drift-detection mechanism.
Forecast error of all model variations were benchmarked in real-time against a state-of-the-art numerical weather prediction model.
Performance was assessed using classical and novel evaluation metrics. Results indicate that using the Variance Horizon reduced computational usage by more than 50\%, while increasing between 0-15\% in error. Meanwhile, performance-based retraining reduced computational usage by up to 90\% while \emph{also} improving forecast error by up to 10\%. Finally, the combination of both the Variance Horizon and performance-based retraining outperformed other model configurations by up to 99.7\% when considering error normalized to computational usage.
\end{abstract}

\newpage
\section{Introduction}
\label{sec:intro}
\vspace{-1pt}
Technological advances have enabled a continuous flow of data from an ever-growing assortment of sources, which is promoting data-driven decision-making as a fundamental component of many aspects of our society. Given the vastness and richness of the available data, machine learning lends itself as a promising tool for decision-making in a variety of fields.
The typical machine learning paradigm operates under the assumption of distribution stationarity, i.e., the data is independently and identically distributed (i.i.d.) \cite{vapnik1999nature}. However, many systems display temporally varying data distributions that violate this stationarity assumption, including meteorology, financial markets, fraud detection, and other real-time applications. 
This distribution shift is called \textit{concept drift} and can lead to deteriorated model performance \cite{gepperth2016incremental}.

A common approach to address this issue is \textit{incremental learning}. Unlike traditional batch learning, where the model is trained on a fixed data set and then deployed, incremental learning is an online learning approach that updates the model parameters continuously as new data becomes available. As such, it allows the model to adapt to changes in the data-generating process and update its predictions accordingly, thus reducing the impact of concept drift \cite{AGRAHARI2022}.
Furthermore, some data sources may evolve such that including data from too far back in time becomes detrimental to the forecasting performance. 
For example, meteorological nowcasting aims to provide hyper-local, short-term, weather forecasts. This short-term horizon means that the model has little need for vast amounts of historical data that may have conflicting representations how the short-term forecast horizon may be changing. 
Consequently, these types of use-cases benefit from limiting the amount of historical data used to train machine learning models and hence come at a lower computational cost. 


The balance between computational cost and prediction error of non-parametric regression techniques are primary concerns shared across many domains \cite{ALJARRAH2015, Laudani2105}. Specifically, a large segment of machine learning techniques for time-series forecasting are geared toward detecting concept drift to help improve model performance while minimizing computational cost \cite{AGRAHARI2022}.
However, few studies make comparisons of error trade-off for computational cost savings. \citet{AHMAD2018} compared random forests and decision trees and found that decision trees are by far the most computationally efficient algorithms, with a sacrifice to error. \citet{Sigtia2016} compared operation counts for support vector machines, deep neural networks (DNNs), and Gaussian mixture models to conclude that DNNs have the best error to cost ratio for sound classification problems. Although these studies yield valuable information, they do not provide insight into common regression techniques for real-time, time-series forecasts. Furthermore, concept drift in time-series and regression tasks is an under-investigated research area \cite{bayram2022concept}.

In an effort to fill these gaps in literature, the present study explores the relationship between forecast error and computational cost, with an objective of minimizing the latter, through an example use-case regression task prone to concept drift: meteorological nowcasting. 
We hypothesize that computational cost can be minimized with little to no sacrifice to prediction error via two methods: a variance controlled training data set and a concept drift detection retrain criteria.
The relationship between forecast error and computational cost is quantified via a real-time parameter sweep that includes 4000 data sources and 144 total model hyperparameter combinations, each one a variation of (1) three meteorological variables: temperature, wind speed, and cloud cover; (2) four different machine learning model architectures: XGBoost boosted decision tree ensemble, fully-connected multi-layer perceptron (FC-MLP), Transformer with positional encoding, and Long Short-Term Memory network (LSTM); (3) two retrain frequencies: hourly and performance-based concept drift detection; and (4) with and without an auto-adaptive training window (the Variance Horizon).
The results are compared to state-of-the-art numerical weather prediction (NWP) models using classical error estimation methods as well as novel scores geared for industrial applications, including error per kilowatt hour [$\varepsilon/ \mathrm{kwH}$].
Experimental results indicate that the proposed approach achieves competitive performance whilst offering the advantage of adaptability to the evolving meteorological data generating process in combination with a significant reduction in computational cost. The findings of this study have the potential to improve the computational effort of real-time adaptive modeling and benefit a wide range of applications that rely on accurate time-series predictions.

\newpage
\subsection{Machine learning for meteorology}
\label{subsec:mlmeteo}

Accurate, rapid, and cost effective forecasting of time-series meteorological variables such as temperature and wind speed is crucial for a wide range of applications, including agriculture, transportation, and disaster management. 
Conventionally, meteorological forecasting is done through NWP models that describe the physical processes that impact the spatio-temporal evolution of the variables of interest \cite{lynch2006emergence}. To produce such forecasts, NWP models require massive simulations to compute the complex dynamics of the Earth's atmospheric system through solving coupled differential equations \cite{bauer2015quiet}. With increased computational power and availability of data sets, data-driven machine learning approaches have become a rapidly evolving area of research to improve accuracy and performance, and provide alternatives to the classical NWP methods \cite{chase2022machine, schultz2021can}. A plethora of model architectures, from decision trees to graph neural networks and Transformers, have been explored for a variety of applications. Examples include predicting humidity at different altitudes \cite{chisholm1968diagnosis}, next-day severe weather forecasting \cite{loken2020rf}, radar retrieval of snowfall \cite{chase2021dual}, estimating tropical cyclone intensity for satellite data \cite{chen2019estimating, griffin2022predicting}, automatic detection of cyclones from satellite images \cite{kumler2020tropical}, global weather forecasting \cite{keisler2022forecasting, lam2022graphcast}, and multi-variable weather forecasting \cite{man2023w}.

Weather nowcasting is a sub-field of meteorology that focuses on hyper-local forecasting at short lead-times, typically up to six hours \cite{schmid2019}. 
From a machine learning perspective, nowcasting can be regarded as a sequence forecasting problem where the dynamic system to be modeled is a spatial region represented by a \(N\times M\) grid. 
Each grid cell constitutes a location that is represented by \(P\) temporally varying measurements. 
As such, the parameter space, \(\mathbf{R}\), at time t has the dimensions \(M \times N \times P\) and can be represented by a tensor \(\mathcal{X}_t\in \mathbf{R}^{M \times N \times P}\). 
Given a sequence of \(J\) previous observations, \(\mathcal{X}_{t-J+1}, \mathcal{X}_{t-J+2}, .., \mathcal{X}_t\), the goal is to predict a sequence \(\mathcal{\widetilde{Y}}_{t+1}, \mathcal{\widetilde{Y}}_{t+2}, .., \mathcal{\widetilde{Y}}_{t+K}\), where \(\mathcal{\widetilde{Y}}\) is a tensor containing the full set or a subset of the variables in \(\mathcal{X}\), and \(K\) is the furthest forecast horizon. 
As for meteorology in general, nowcasting has garnered the attention of the machine learning community \cite{kuligowski1998experiments, shi2015convolutional, wang2020rf, stock2021using, samsi2019distributed}.
However, whilst literature demonstrates a plethora of examples where machine learning has been used for nowcasting applications, most works rely on historical data; end-to-end implementations where real-time data is used are more sparse \cite{verma2020real, fowdur2022real}.

While literature shows a clear increase in machine learning applications for meteorology use-cases, NWP models still prevail as state-of-the-art. Two core arguments for the resistance toward machine learning techniques are the lack of explainability and the lack of physical constraints associated with typical machine learning implementations \cite{schultz2021can}. It is clear that further research and development is necessary for proper adoption of machine learning in weather prediction. 
To this end, software frameworks allowing for systematic and reproducible studies are paramount. The present paper addresses this by deploying an open-source, Python-based, software framework - Flowdapt - designed specifically for optimizing systematic and efficient data handling through construction of computational graphs to avoid duplication and guarantee reproducibility.

\section{Methodology}
\label{sec:methods}

The hypotheses presented in Section \ref{sec:intro} were tested by conducting a live experiment geared toward the collection of prediction error and computational efficiency for the set of 144 model parameter combinations shown in Table~\ref{tab:experimental_design}. These parameter combinations include four different machine learning model architectures (XGBoost, FC-MLP, Transformer, LSTM) used for multi-horizon forecasting of three meteorological variables (temperature, wind speed, and cloud cover) for three cities (Los Angeles, Miami, and Boston). All combinations were compared using hourly retraining, performance based retraining (Section \ref{subsec:cc}), as well as an adaptive training window (Variance Horizon Section \ref{subsec:dataset}). The experiment was run for 110 hours, between the 12th and 17th of May, 2023. 

\begin{table}[htbp]
    \centering
    \caption{Experimental design comprising 144 parametric combinations.
    \label{tab:experimental_design}}
    \begin{tabular}{r|cccc}
        \toprule
        \emph{Training frequency} & \multicolumn{2}{c}{Hourly} & \multicolumn{2}{c}{Performance based} \\
        \emph{Data selection} & \multicolumn{2}{c}{Full set} & \multicolumn{2}{c}{Variance Horizon} \\
        \emph{City} & \multicolumn{1}{c}{Boston} & \multicolumn{2}{c}{Los Angeles} & \multicolumn{1}{c}{Miami}\\
        \emph{Target} & \multicolumn{1}{c}{Temperature} & \multicolumn{2}{c}{Wind speed} & \multicolumn{1}{c}{Cloud cover}\\
        \emph{Regressor} & \multicolumn{1}{c}{XGBoost} & \multicolumn{1}{c}{FC-MLP} & \multicolumn{1}{c}{Transformer} & \multicolumn{1}{c}{LSTM} \\
        \bottomrule
    \end{tabular}
\end{table}

\subsection{Adaptive training strategy}
\label{sec:adaptive_retrain}
The study employed a version of incremental learning where a new minimal model is periodically retrained from scratch using a sliding window through time. This retraining technique enables rapid adaptivity to chaotic systems, and has already seen success in finance applications \cite{Caulk2022}. At its core, the approach fits a minimal functional description to a hyper-local parameter space such that the trained model is only useful for the forecast horizon, beyond which the functional description is no longer applicable. This strategy avoids the potential negative impact of incorporating historical data which may be irrelevant to the present short-term forecast. 

\subsection{Data set}
\label{subsec:dataset}

Raw data consisted of hourly time-series features collected from the open-source OpenMeteo API~\cite{openmeteo-docs}. The API generates data using ensemble forecasting, a technique that combines multiple state-of-the-art NWP forecasts to reduce uncertainties in the data. 
Specifically, the obtained data was a combination of forecasts generated by the Global Forecast System (GFS) and the High-Resolution Rapid Refresh (HRRR), both operated by the National Oceanic and Atmospheric Administration of the USA (NOAA) \cite{Zippenfenig_2022_2, derber1989global, tang2022evaluation, gfs, dowell2022high, james2022high}. 
Using the OpenMeteo \textit{GFS Seamless} option, which automatically selects the best option of the GFS and HRRR forecasts for the specific spatial location \cite{Zippenfenig_2022},
27 base features (listed in Supplementary Table \ref{sup-tab:variables}) were collected for 49 grid points in equidistant grids of size 300 km\(\times\)300 km centered on three USA cities: Los Angeles, Miami, and Boston (see Supplementary Figure \ref{sup-fig:map}). For each grid point, 150-days of historical data was obtained. Consequently, the data set for each of the three cities included historical features from the corresponding grid points plus two additional features for temporal encoding using the sine and cosine transforms of the hour-of-day as additional variables. 

\paragraph{Data preprocessing}

Features containing more than 40 missing values were dropped from the data set. Likewise, time points containing missing data were dropped. The remaining features were filtered using a variance threshold to further remove any static features. Finally, the data set was split into training and validation data (see further details below). The training data and labels were subsequently scaled to [0, 1] using min-max normalization; the same transformation was applied to the validation features and validation labels.

\paragraph{Training horizon}
Two methods for choosing the required amount of training data were investigated:

\textit{Static window} \hspace{5pt} The 150-day data set was split such that the 10 most recent days were used for validation and the remaining data was used for training.

\begin{wrapfigure}{r}{0.52\textwidth} 
\vspace{-17pt}
  \begin{center}
    \includegraphics[width=0.5\textwidth]{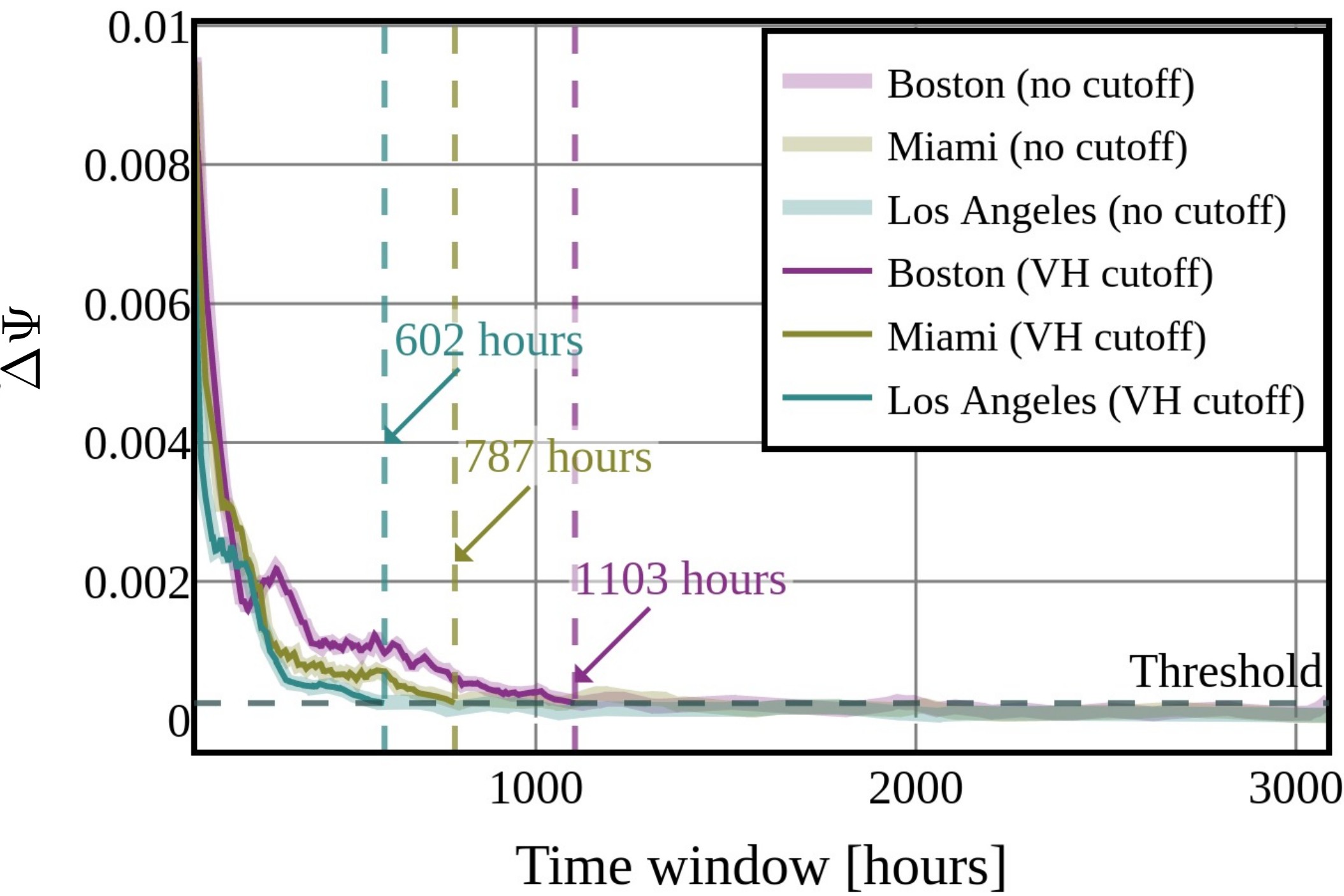}
    \vspace{-5pt}
    \caption{Example of auto-adaptive training window determination using the Variance Horizon. These representative curves showcase how the Variance Horizon cutoff significantly reduces the amount of historical data needed to train the models.
    \label{fig:variance-horizon}}
  \end{center}
\end{wrapfigure}

\textit{Auto-adaptive window} \hspace{5pt} In an effort to reduce the computational cost, we propose a new method for determining the size of the training data set - the Variance Horizon (\(t_{\mathrm{vh}}\)). The Variance Horizon, \(t_{\mathrm{vh}}\), is based on measuring the variance of the pairwise Euclidean distances, \(\textrm{d}\), in the forecast horizon (six hours in the present nowcasting example) and comparing this variance to the variance of distances in a growing training window (up to 3600 hours in the present study). We define the forecast horizon as \(\mathbf{x}\) and the growing training window as \(\mathbf{y}_t\), with the subscript \(t\) indicating the number of points to include. The pairwise distances between the forecast horizon and the training window follows:

\begin{equation}
\mathbf{D}_{xy}(t) = \textrm{d}(\mathbf{x}, \mathbf{y}_t)
\end{equation}

The variances of these distances explains the relative variance between our forecast horizon and the training horizon. Similarly, the pairwise distances within the growing training window explains the total variance explained by the training data:

\begin{equation}
\mathbf{D}_y(t) = \textrm{d}(\mathbf{y}_t)
\end{equation}

\noindent We aim to find the horizon by which the forecast horizon variance is sufficiently explained by the training horizon. Thus, we compute the variance of these two pairwise distances, \(\mathrm{std}(\mathbf{D}_{xy}(t))\) and \(\mathrm{std}(\mathbf{D}_y(t))\) for a growing training window:

\begin{equation}
\label{eq:var-pwd}
    \Psi(t) = \frac{\mathrm{std}(\mathbf{D}_{xy}(t))}{\mathrm{std}(\mathbf{D}_y(t))}
\end{equation}

\noindent Finally, we define a plateau stopping criterion for the iterative solution of \(t_{vh}\):

\begin{equation}
\label{eq:delta-var-pwd}
    \Delta\Psi = \frac{1}{\alpha} \sum^{t_{vh}}_{t_{vh}-\alpha} \frac{\partial \Psi}{\partial t} \leq 0.25\cdot10^{-3} 
\end{equation}

where \(\alpha\) is number of points in the training window within which \(\Delta\Psi\) is to plateau, and the threshold of \(0.25\cdot10^{-3}\) was selected empirically based on the \(t_{\mathrm{vh}}\) example estimates shown in Figure \ref{fig:variance-horizon}.

\paragraph{Feature and label engineering}

A look-back sequence of 24 hours was used as input for all model types. As such, the input dimensions for the Transformer and LSTM consisted of tensors sized [24, 1325]. Meanwhile, the same tensor was flattened to a one-dimensional vector of [1, 31800] for the input to the XGBoost and FC-MLP models. Labels for all four model types were identical: a sequence of the 6 subsequent hours for each of the three forecast variables.

\subsection{Regression methods}

The present paper compares the performance of four regression methodologies in real-time adaptive modeling environments, including XGBoost gradient boosted decision trees, Fully-Connected Multi-Layer-Perceptron (FC-MLP), Transformer with positional encoding, and Long Short-Term Memory network (LSTM). The input features to all four models is identical in dimension, with sequence of 24 hours and each hour containing 1350 features. Furthermore, the three neural network architectures all feed the same base FC-MLP decoder architecture.
Mean Squared Error (MSE) was used as loss function for all models. All ANN architectures were trained using batches of 64 samples, generated by shuffling the training time steps. The commonly used Adam algorithm \cite{kingma2014adam} was used as optimizer, with a learning rate of 5e-5 that was allowed to reduce by half when the validation loss plateaued. Early stopping, with a patience of 20 epochs, was used to speed up the training process. Validation loss was tracked and used to select the best model.

\paragraph{Gradient boosted decision tree}

The implementation of decision trees was done using the open-source Python implementation of XGBoost \cite{xgboost}. Each tree ensemble was allowed to train 40 trees, using a maximum depth of 6 with at least 5 samples per leaf. The L1 regularization term was set to 0.5 and the learning rate was set to 0.1. 

\paragraph{Fully Connected Multi-Layer Perceptron}\label{para:fcmlp}

The implementation of the FC-MLP \cite{mcclelland1987parallel} was done using the open-source PyTorch library \cite{pytorch}. The input layer is sized according to the flattened tensor of [24, 1400]. Three hidden layers, each with 4096, 2056, and 1024 nodes, respectively, were combined with a ReLU activation function to produce an output sequence of \(6\times1\) values corresponding to the 1-6 hour forecasts. To avoid over-fitting, drop-outs \cite{srivastava2014dropout} of 10\% were added to each hidden layer. 

\paragraph{Transformer with positional encoding}

The transformer architecture was implemented in PyTorch using a set of three modules: a dense input layer, designed to ensure the Transformer input was divisible by the number of attention heads, connected to two Transformer encoder layers with positional encoding \cite{vaswani2017attention} and ca. [24, 1400] input dimension, followed by a FC-MLP decoder with the same architectural dimensions as the FC-MLP in the previous paragraph.

\paragraph{Long Short-Term Memory network}

PyTorch was used also for the LSTM architecture, which was a set of 2 modules: an LSTM with 2 layers of ca. [24, 1400] input, followed by a FC-MLP decoder with the same architecture dimensions as the previously mentioned FC-MLP architecture.

\subsection{Computing platform}
\label{subsec:platform}

Real-time adaptive modeling in data-streaming environments presents a variety of problems including API rate limitations, data preparation choke-points, and disk IO read speed choke-points. To address these challenges, we constructed and employed an open-source, workflow software - Flowdapt - designed to relieve these constraints in data-streaming environments. The software schedules individual tasks, such as data fetching, data preparation, model training, inference, and event triggers (such as concept drift detection described in Section \ref{subsec:cc}). These tasks are parallelized and distributed across a cluster using a Dask backend \cite{dask}. Meanwhile, the software adds methods optimized for real-time data sharing across large numbers of data sources and model variations. For example, the software leverages computational graph construction via Dask such that a single set of raw data is stored once to cluster memory, but linked via pointers through a variety of data sets and efficiently manipulated via Dask graph submission. All software is documented and available in the Supplementary Material.

\paragraph{Hardware}

The experiment was powered entirely by solar power on two 16 core 3.9 GHz AMD EPYC 7343 Milan processors, three A4500 GPUs, and 256 GB DDR4 3200 MHz RAM, running on Ubuntu 22.04.

\subsection{Evaluation metrics}

\paragraph{Error versus computational usage}

Model performance relative to the computational usage was quantified by tracking the minimum validation loss and wall time for every trained model (1728/day, or ca. 7,929 models for the entire experiment). Wall time was converted to expended power (kWh) via the power rating of the GPU (A4500, 200W). Train time was used throughout as a proxy for power expended and computational efficiency.

\paragraph{Concept drift detection}
\label{subsec:cc}

During the experiment, the concept drift of each model was evaluated once per hour by reconstructing a new validation data set on a sliding window of 240 hours. The model was inferenced using the new validation data set and the validation loss was compared to the minimum validation loss obtained during the initial model training. Concept drift was defined as a \(\ge\)5\% increase in validation loss for new inference data compared to the original validation loss, upon which a new model was trained to replace the original.

\paragraph{Prediction error} 

Predictions generated by the machine learning models were compared to those obtained from the OpenMeteo GFS Seamless forecast using the root-mean-squared-error (RMSE) metric. Mean and standard deviation per city and model architecture were computed using bootstrapping, where the sample data was a vector of the RMSE values for each of the six forecast horizons.

\section{Results and Discussion}
\label{sec:results}

The relationship between model performance and computational cost was examined using a variety methods for all combinations of model architectures, cities, and meteorological variables. 

\paragraph{Model performance versus computational cost}

\begin{figure}[h!]
    \centering
    \includegraphics{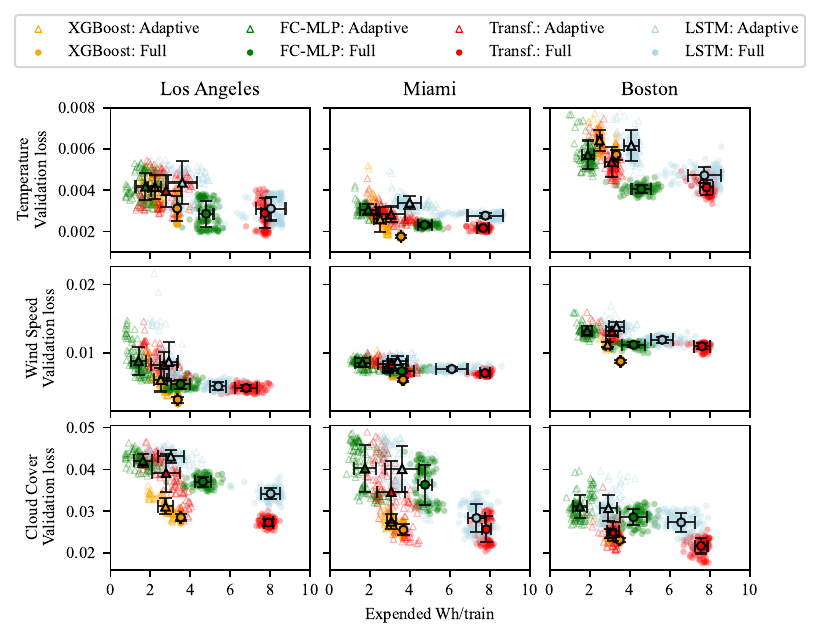}
    \vspace{-5pt}
    \caption{Validations loss vs. computational usage based on training on the full training set (filled circles) compared to using the adaptive training window (empty triangles). Columns show the individual cities (Los Angeles, Miami, and Boston) and rows the meteorological variables (temperature, wind speed, and cloud cover) for each model architecture (XGBoost - orange, FC-MLP - green, Transformer - red, LSTM - light blue).
    \label{fig:power_expended_valloss}}
\vspace{-17pt}
\end{figure}

As shown in Figure \ref{fig:power_expended_valloss}, the mean validation loss of all 7,290 trained models tended to decrease as expended power increased (tabular results also available in Supplementary Tables \ref{sup-tab:val-pwr-temp}-\ref{sup-tab:val-pwr-cc}). Furthermore, Figure \ref{fig:power_expended_valloss} shows that the Transformer trained with the full 150-day data set was the most power consuming combination, achieving the lowest validation loss in 5/9 cases. However, XGBoost with a full data set used approximately half the power while reaching equivalent or even lower validation loss in 4/5 cases. 

\begin{wrapfigure}{r}{0.5\textwidth} 
\vspace{-25pt}
  \begin{center}
    \includegraphics[width=0.48\textwidth]{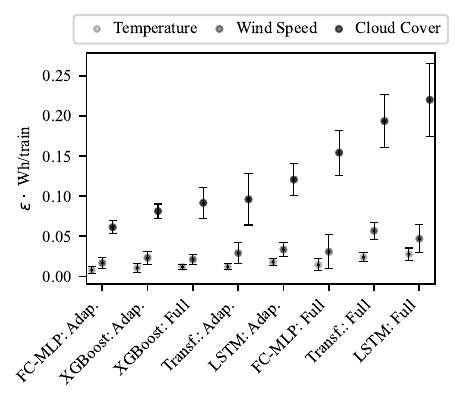}
\vspace{-5pt}
    \caption{Minimum validation loss normalized by power expended (Wh) for each model configuration with (Adapt.) and without (Full) the adaptive training window obtained through using the Variance Horizon, averaged over all three cities.
    \label{fig:norm_perf}}
  \end{center}
\vspace{-17pt}
\end{wrapfigure}

The mean validation loss vs. computational cost shown in Figure \ref{fig:power_expended_valloss} was averaged across cities to summarize the cost-normalized effect of the Variance Horizon (\(t_{\mathrm{vh}}\)) in Figure \ref{fig:norm_perf}, where the adaptive training window generally improved the cost-normalized performance of all architectures.

Specifically, the use of the adaptive training window improved FC-MLP by 65\%. Figure \ref{fig:power_expended_valloss} also shows that XGBoost was the model architecture least affected by the amount of training data.
In regard to the meteorological variables, cloud cover proved to be more difficult for all model architectures to capture than the temperature and wind speed, regardless of amount of training data. One possible explanation for this is the dissimilarities in temporal distribution: while temperature and wind speed showed smoothly varying values throughout the duration of the experiment, cloud cover had a more binary behavior jumping between 0\% and 100\% with few intermittent values (see Supplementary Figures \ref{sup-fig:temp-la}-\ref{sup-fig:cc-bos}).

\begin{figure}[h!]
    \centering
    \includegraphics{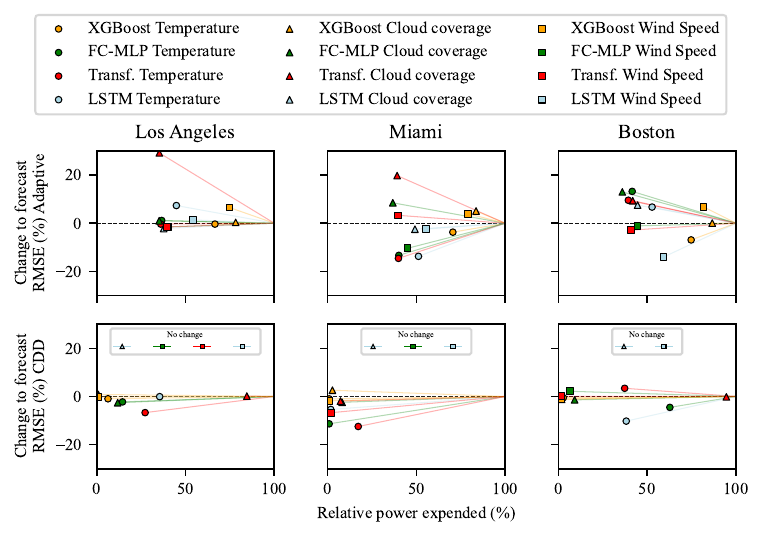}
    \caption{Effect of the Adaptive training window (using Equation \ref{eq:delta-var-pwd}) and Concept Drift Detection (CDD) on real-time forecast error (RMSE) and power expended RMSE values for each model and target are computed for all 110 predictions made during the 110-hour experiment.
    \label{fig:power_expended_rmse}}
\end{figure}


The real-time forecast RMSEs for all Los Angeles model combinations shown in Table~\ref{tab:la-rmse} indicate that the temperature prediction error (RMSE) for all models versus ground truth were approximately 1.35 \degree C (Tables for Boston and Miami are available in Supplementary Table \ref{sup-tab:rmse}). Meanwhile, the comparison between regressors and the GFS benchmark showed that regressor RMSE ranged from equivalent to 10\% worse error. Wind speed and cloud cover prediction error-trends also confirmed that the regressor performances were equivalent, or in many cases superior, to the GFS benchmark. 

Compared to the ground truth, all models showed an RMSE of approx. 2.3 km/h and 35\% for wind-speed and cloud cover, respectively. In all cases, the models were well within the large gaps observed between major NWP models such as GEM (run by the Canadian Meteorological Center), which had worse RMSEs relative to the GFS historical data of 2.4\(\degree\)C, 59.0\%, and 4.5 kph for the temperature, cloud cover, and wind speed, respectively, during the same period as the current experiment.

\paragraph{Effect of concept drift detection and adaptive training window}
As shown in Figure \ref{fig:power_expended_rmse}, both of the primary parameters of the present study (concept drift detection (CDD) and the adaptive training window) played important roles on the final real-time forecast error and expended power. In both cases, the power expended decreased between 20\% and nearly 99\%. It is clear that the adaptive training window reduced expended power by increasing forecast error by 5-15\% for 15/36 models. The remaining 21 models reached equivalent or better RMSE with a reduced power consumption of ca.~50\%. Meanwhile, CDD improved RMSE for 24/36 models by up to 10\% and reducing expended power by ca.~75\%. Of the remaining 12 CDD models, the maximum increase of RMSE was less than 4\%. Supplementary Tables \ref{sup-tab:val-pwr-temp}-\ref{sup-tab:val-pwr-cc} show the variance for the expended power observations as merely 5-10\%, while Table~\ref{tab:la-rmse} shows how the forecast RMSE observations varied by as little as 1-5\%.

Figure \ref{fig:power_expended_rmse} also shows how cloud cover did not benefit from the adaptive training window or the CDD. In fact, drift was detected for all cloud cover data points resulting in hourly retraining. Further, the cloud cover forecasts were among the worst for the adaptive training windows. Despite these limitations, the cloud cover forecasts were still superior to the GFS benchmark forecasts for 2/3 cities (see Tables \ref{tab:la-rmse} and \ref{sup-tab:rmse}).

The primary reason that CDD improved forecast error can be explained by it preferencing the best models as opposed to the naive hourly retraining that simply takes a new model each hour whether the validation loss has improved or not relative to the previous models. Additionally, CDD allows for adapting to changing parameter spaces when necessary, while avoiding retraining when the parameter space remains constant. This is especially important for unpredictable and chaotic systems, such as meteorology, where certain parameter spaces for targets like cloud cover may change radically for one location and require retraining, but the wind speed parameter space may stay relatively constant and require much less frequent retraining. These evolving parameter spaces are confirmed by tracking the validation loss with time as shown in Supplementary Figure \ref{sup-fig:loss_with_time}. Thus, CDD allows for an increased level of adaptivity when deployed across a diverse range of chaotic targets. Finally, the CDD methodology (Section \ref{subsec:cc}) chosen for the present nowcasting problem was straightforwardly implemented since the validation data set was easily labeled and updated upon arrival of new data points.

Finally, Table~\ref{tab:la-rmse} shows that combining the adaptive training window with CDD yields an improved computational cost-normalized performance of 99.7\% for XGBoost compared to the LSTM when retraining hourly on the full data set (cloud cover RMSE\(\cdot\)Wh/prediction of 0.87\% vs. 329.8\%). For XGBoost, validation loss for new data was less often above the 5\% validation loss threshold and so the model needed retraining less often, resulting in lower computational cost. As such, XGBoost appeared to be better at capturing the behavior of the data. Decision trees are well known for capturing trends in heterogeneous parameter spaces due to their ensemble modeling approach to fitting features from different statistical distributions (e.g., precipitation and temperature).

An important comparison of power consumption between the state-of-the art NWP (GFS) and the regression techniques here was not possible for multiple reasons. First, NOAA does not publish the computational cost of their models, and second, their models are run for dense grids across the entire globe/USA. This means that even if NOAA did publish computational usage, it would not be possible to determine the computational cost required to make a hyper-local, short-term nowcast at a specific coordinate, as is done with the regression techniques in the current paper. While these comparison limitations are clear, it is important to note that the GFS model is running on National Laboratory supercomputers, while the regressors here are running on a single solar-powered server.

\begin{table}[h!]
    \centering
    \caption{Prediction error (RMSE) with and without computational cost normalization for Los Angeles. Mean \(\pm\) std (for all prediction horizons) model-specific performance for the different combinations of training horizon (full versus adaptive window) and retraining frequency (hourly versus using concept drift detection (CDD)). 
    \textbf{Bold} values indicate the optimal combination of training horizon, retraining frequency, and model for each individual variable.
    \label{tab:la-rmse}}
    \adjustbox{max width=\textwidth}{%
    \begin{tabular}{@{}lcccc|cccc@{}}
        \toprule
        & & \multicolumn{6}{c}{Temperature, \degree C} \\
        \cmidrule(lr){2-9}
        & \multicolumn{4}{c}{RMSE} & \multicolumn{4}{c}{RMSE\(\cdot\)Wh/prediction} \\
        \cmidrule(lr){2-5} \cmidrule(lr){6-9}
        & \multicolumn{2}{c}{Full} & \multicolumn{2}{c}{Adaptive} & \multicolumn{2}{c}{Full} & \multicolumn{2}{c}{Adaptive}\\
        \cmidrule(lr){2-3} \cmidrule(lr){4-5} \cmidrule(lr){6-7} \cmidrule(lr){8-9} 
        Model & Hourly & CDD & Hourly & CDD & Hourly & CDD & Hourly & CDD \\
        \midrule \\[-1.3ex]
        XGBoost 
        & 1.38\(\pm\)0.12 & 1.36\(\pm\)0.12 & 1.38\(\pm\)0.12 & 1.4\(\pm\)0.13
        & 4.61\(\pm\)0.42 & 0.29\(\pm\)0.02 & 3.06\(\pm\)0.25 & \textbf{0.17\(\pm\)0.02} \\
        FC-MLP 
        & 1.41\(\pm\)0.13 & 1.38\(\pm\)0.12 & 1.43\(\pm\)0.13 & 1.43\(\pm\)0.13
        & 6.78\(\pm\)0.62 & 0.96\(\pm\)0.08 & 2.51\(\pm\)0.23 & 1.05\(\pm\)0.1 \\
        Transformer 
        & 1.4\(\pm\)0.12 & \textbf{1.3\(\pm\)0.11} & 1.39\(\pm\)0.13 & 1.34\(\pm\)0.12
        & 10.82\(\pm\)0.93 & 2.75\(\pm\)0.23 & 3.89\(\pm\)0.37 & 0.54\(\pm\)0.05 \\
        LSTM 
        & 1.37\(\pm\)0.12 & 1.37\(\pm\)0.12 & 1.47\(\pm\)0.14 & 1.43\(\pm\)0.13
        & 11.04\(\pm\)1.01 & 3.91\(\pm\)0.36 & 5.32\(\pm\)0.52 & 3.1\(\pm\)0.29 \\
        GFS 
        & \multicolumn{4}{c}{1.33\(\pm\)0.11}
        & \multicolumn{4}{c}{-}\\        
    \end{tabular}%
    }
    \adjustbox{max width=\textwidth}{%
    \begin{tabular}{@{}lcccc|cccc@{}}
        \toprule
        & & \multicolumn{6}{c}{Wind speed, km/h} \\
        \cmidrule(lr){2-9}
        & \multicolumn{4}{c}{RMSE} & \multicolumn{4}{c}{RMSE\(\cdot\)Wh/prediction} \\
        \cmidrule(lr){2-5} \cmidrule(lr){6-9}
        & \multicolumn{2}{c}{Full} & \multicolumn{2}{c}{Adaptive} & \multicolumn{2}{c}{Full} & \multicolumn{2}{c}{Adaptive}\\
        \cmidrule(lr){2-3} \cmidrule(lr){4-5} \cmidrule(lr){6-7} \cmidrule(lr){8-9} 
        Model & Hourly & CDD & Hourly & CDD & Hourly & CDD & Hourly & CDD \\
        \midrule \\[-1.3ex]
        XGBoost 
        & 1.92\(\pm\)0.03 & \textbf{1.92\(\pm\)0.02} & 2.05\(\pm\)0.06 & 2.13\(\pm\)0.07
        & 6.5\(\pm\)0.11 & 0.06\(\pm\)0.0 & 5.17\(\pm\)0.14 & \textbf{0.05\(\pm\)0.0} \\
        FC-MLP 
        & 2.44\(\pm\)0.09 & 2.44\(\pm\)0.09 & 2.4\(\pm\)0.07 & 2.35\(\pm\)0.06
        & 8.52\(\pm\)0.32 & 8.52\(\pm\)0.32 & 3.38\(\pm\)0.1 & 0.72\(\pm\)0.02\\
        Transformer 
        & 2.46\(\pm\)0.09 & 2.46\(\pm\)0.09 & 2.41\(\pm\)0.06 & 2.4\(\pm\)0.06
        & 16.63\(\pm\)0.62 & 16.63\(\pm\)0.62 & 6.49\(\pm\)0.17 & 0.47\(\pm\)0.01 \\
        LSTM 
        & 2.47\(\pm\)0.08 & 2.47\(\pm\)0.08 & 2.5\(\pm\)0.1 & 2.35\(\pm\)0.07
        & 13.27\(\pm\)0.45 & 13.27\(\pm\)0.45 & 7.28\(\pm\)0.31 & 3.18\(\pm\)0.1 \\
        GFS 
        & \multicolumn{4}{c}{2.07\(\pm\)0.03}
        & \multicolumn{4}{c}{-}\\        
    \end{tabular}%
    }
    \adjustbox{max width=\textwidth}{%
    \begin{tabular}{@{}lcccc|cccc@{}}
        \toprule
        & & \multicolumn{6}{c}{Cloud cover, \%} \\
        \cmidrule(lr){2-9}
        & \multicolumn{4}{c}{RMSE} & \multicolumn{4}{c}{RMSE\(\cdot\)Wh/prediction} \\
        \cmidrule(lr){2-5} \cmidrule(lr){6-9}
        & \multicolumn{2}{c}{Full} & \multicolumn{2}{c}{Adaptive} & \multicolumn{2}{c}{Full} & \multicolumn{2}{c}{Adaptive}\\
        \cmidrule(lr){2-3} \cmidrule(lr){4-5} \cmidrule(lr){6-7} \cmidrule(lr){8-9} 
        Model & Hourly & CDD & Hourly & CDD & Hourly & CDD & Hourly & CDD \\
        \midrule \\[-1.3ex]
        XGBoost 
        & 32.78\(\pm\)1.23 & 33.06\(\pm\)1.22 & 32.88\(\pm\)1.22 & 34.38\(\pm\)1.0
        & 116.02\(\pm\)4.43 & 1.06\(\pm\)0.04 & 91.09\(\pm\)3.42 & \textbf{0.87\(\pm\)0.03} \\
        FC-MLP 
        & 38.86\(\pm\)1.27 & 37.87\(\pm\)0.96 & 39.3\(\pm\)1.06 & 39.12\(\pm\)1.07
        & 180.69\(\pm\)6.01 & 20.8\(\pm\)0.55 & 64.29\(\pm\)1.73 & 23.3\(\pm\)0.65 \\
        Transformer 
        & \textbf{30.52\(\pm\)0.89} & 30.57\(\pm\)0.89 & 39.4\(\pm\)1.11 & 38.62\(\pm\)1.03
        & 242.03\(\pm\)7.21 & 204.98\(\pm\)6.16 & 110.35\(\pm\)3.12 & 6.89\(\pm\)0.19 \\
        LSTM 
        & 40.98\(\pm\)1.68 & 40.98\(\pm\)1.68 & 40.12\(\pm\)1.24 & 39.69\(\pm\)1.19
        & 329.81\(\pm\)13.65 & 329.81\(\pm\)13.65 & 121.72\(\pm\)3.76 & 33.98\(\pm\)1.04  \\
        GFS 
        & \multicolumn{4}{c}{33.88\(\pm\)0.76}
        & \multicolumn{4}{c}{-}\\        
    \end{tabular}%
    }
\end{table}

\section{Conclusion}
\label{sec:conclusion}

This paper presents an in-depth study of the relationship between forecast error and computational cost for a time-series, regression task where the data suffers from concept drift. The results from a real-time, parametric experiment show that nowcasting (hyper-local, short-term forecasting) can be efficiently achieved through machine learning techniques that rival state-of-the-art numerical methods in terms of prediction error while offering reduced computational costs. This improved performance is made possible by the advent of the Variance Horizon and further improved by concept drift detection-based retraining of models.

The findings of this study have the potential to improve the computational effort of real-time adaptive modeling and benefit a wide range of applications that rely on accurate time-series predictions. Specifically, the results emphasize the opportunity of using both a reduced data set size as well as a performance based retraining method for nowcasting use-cases. Both of these methods reduce power expenditure while maintaining competitive prediction performance.

\bibliographystyle{unsrtnat}
\small
\setlength{\bibsep}{0pt plus 0.3ex}
\bibliography{references.bib}

\newpage

\renewcommand{\figurename}{Supplementary Figure}
\renewcommand{\thefigure}{S\arabic{figure}}
\renewcommand{\tablename}{Supplementary Table}
\renewcommand{\thetable}{S\arabic{table}}

\section*{Supplementary Material}
\label{sec:sup-material}

\subsection*{Data set}
\label{subsec:sup-dataset}

\begin{figure}[h]
    \centering
    \includegraphics[width=0.8\linewidth]{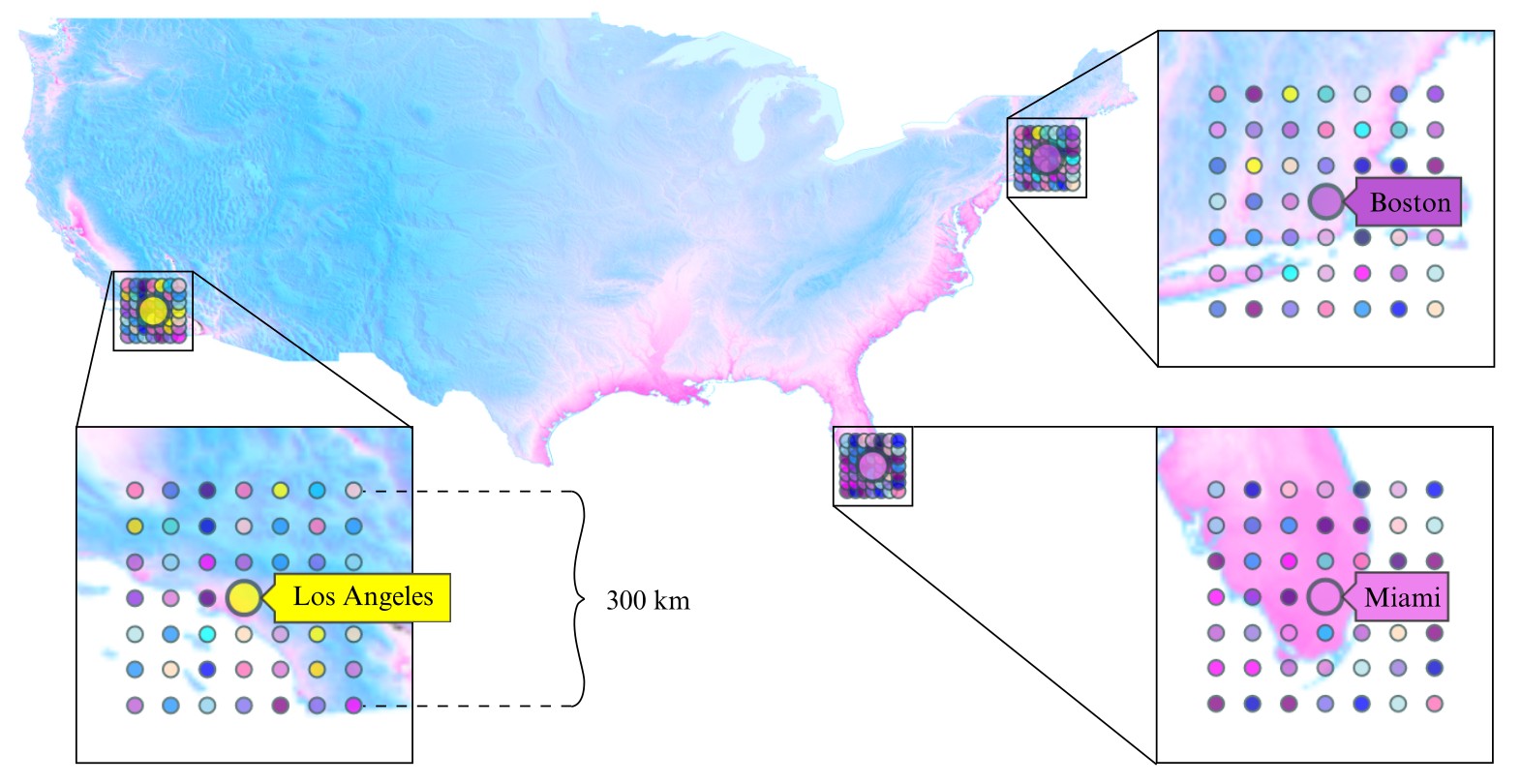}
    \caption{City and grid locations. Meteorological data was acquired for locations on equidistant grids of 7\(\times\)7 points, covering 300\(\times\)300 km, centered around each of the cities Los Angeles, Miami, and Boston.
    \label{sup-fig:map}}
\end{figure}

Multi-horizon (1hr-6hr) forecasting of temperature, wind speed, and cloud cover was done for three cities (Los Angeles, Miami, and Boston) using time series data obtained for equidistant 300\(\times\)300 km grids centered on each city (Supplementary Figure \ref{sup-fig:map}). Raw data consisted of hourly time-series features as per Supplementary Table \ref{sup-tab:variables}.

\begin{table}[ht!]
\footnotesize
    \caption{Meteorological variables used as base features. Variables in \textbf{bold} were used as forecast labels. See \url{https://open-meteo.com/en/docs} for details.
    \label{sup-tab:variables}}
    \begin{tabular}{| p{3.9cm} | p{1.5cm} | p{7.3cm} |}
        \hline
        Variable name & Unit & Description\\
        \hline
        \textbf{Temperature (2 m)} & \degree C & Air temperature at 2 m above ground \\
        Apparent temperature & \degree C & Perceived feels-like temperature \\
        \textbf{Cloud cover} & \% & Total cloud cover as an area fraction \\
        Cloud cover (low) & \% & Low level clouds and fog up to 3 km altitude \\
        Cloud cover (mid) & \% & Mid level clouds from 3 to 8 km altitude \\
        Cloud cover (high) & \% & High level clouds from 8 km altitude \\
        \textbf{Wind speed (10 m)} & km/h & Wind speed at 10 m above ground \\
        Wind speed (80 m) & km/h & Wind speed at 80 m above ground \\
        Wind direction (10 m) & \degree & Wind direction at 10 m above ground \\
        Wind direction (80 m) & \degree & Wind direction at 80 m above ground \\
        Wind gust (10 m) & km/h & Gusts at 10 m above ground (max of preceding hour) \\
        Rel. humidity (2 m) & \% & Relative humidity at 2 m above ground \\
        Dew point (2 m) & \degree C & Dew point temperature at 2 m above ground \\
        Vapor pressure deficit & kPa & Difference of air moisture content from full saturation \\
        Ref. evapotranspiration & mm & Reference emission of water vapor from soil and plants \\
        Precipitation & mm & Total precipitation (sum of prev. hour) \\
        Freezing-level height & m & Altitude above sea level of the 0\degree C level \\
        Weather code & WMO code & Weather interpretation codes \\
        Cape & J/kg & Convective available potential energy \\
        Short-wave radiation & W/m\(^2\) & Shortwave solar radiation (prev. hour avg.) \\
        Short-wave radiation (instant) & W/m\(^2\) & Shortwave solar radiation (at current time) \\
        Dir. radiation & W/m\(^2\) & Direct solar radiation on horizontal plane (prev. hour avg.) \\
        Dir. radiation (instant) & W/m\(^2\) & Direct solar radiation on horizontal plane (at current time) \\
        Dir. normal irradiance (instant) & W/m\(^2\) & Direct solar radiation on normal plane (at current time) \\
        Dif. radiation & W/m\(^2\) & Diffuse solar radiation (prev. hour avg.) \\
        Dif. radiation (instant) & W/m\(^2\) & Diffuse solar radiation (at current time) \\
        Ter. radiation (instant) & W/m\(^2\) & Terrestrial solar radiation (at current time) \\
        \hline
    \end{tabular}
\end{table}

\subsection*{Concept drift detection}
\label{subsec:sup-cc}

The rate of concept drift was assessed through the minimum validation loss of the hourly retrained model with time. As shown in Supplementary Figure \ref{sup-fig:loss_with_time}, the minimum validation loss for cloud cover in Miami changed up to 60\% during a period of 100 hours, indicating the need for rapid model retraining. Meanwhile, the validation loss for wind speed in Los Angeles stayed constant for the same period of time. These rates support the live measured forecasting accuracy, where the constant retraining models performed best on the variables that had the greatest rate of change (see Supplementary Table \ref{sup-tab:rmse}).

\begin{figure}[h!]\centering
     \includegraphics[width=\textwidth]{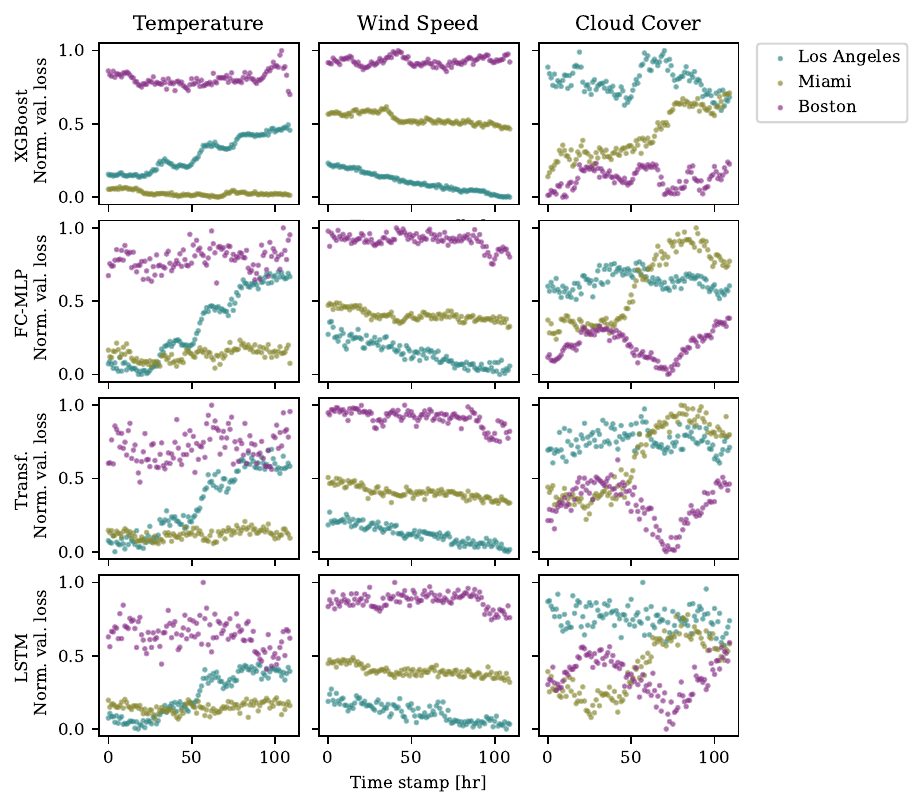}
    \caption{Evolution of minimum validation loss with time. The validation loss was normalized per model for readability, using the min-max values for all cities.
    \label{sup-fig:loss_with_time}}
\end{figure}

\subsection*{Results}
\label{sec:sup-results}

Prediction accuracy (RMSE compared to ground truth), with and without electrical cost normalization, for Miami and Boston is shown in Supplementary Table \ref{sup-tab:rmse}; data for Los Angeles is shown in Table 2 in the manuscript.
Validation loss and expended power for each model, using the full or adaptive training data set, is shown in Supplementary Tables \ref{sup-tab:val-pwr-temp}-\ref{sup-tab:val-pwr-cc}.

Forecasts compared to ground truth for each of the meteorological variables (temperature, wind speed, and cloud cover) and model architecture (XGBoost, FC-MLP, Transformer, and LSTM), as well as the reference NWP model GFS, per city (Los Angeles, Miami, and Boston) when using hourly retraining on the full data set are seen in Supplementary Figures \ref{sup-fig:temp-la}-\ref{sup-fig:cc-bos}.

\begin{table}[h!]
    \centering
    \caption{Prediction accuracy (RMSE) with and without electrical cost normalization. Mean \(\pm\) std (for all prediction horizons) model-specific performance for the different combinations of training horizon (full versus adaptive window) and retraining frequency (hourly versus using concept drift detection (CDD)). 
    \textbf{Bold} values indicate the optimal combination of training horizon, retraining frequency, and model for each individual variable.
    \textbf{\textit{Bold italicized}} GFS values indicate higher accuracy than any machine learning model.
    \label{sup-tab:rmse}}
    \adjustbox{max width=\textwidth}{%
    \begin{tabular}{@{}lcccc|cccc@{}}
        \toprule
        \large \textbf{Miami} & & \multicolumn{6}{c}{Temperature, \degree C} \\
        \cmidrule(lr){2-9}
        & \multicolumn{4}{c}{RMSE} & \multicolumn{4}{c}{RMSE\(\cdot\)Wh/prediction} \\
        \cmidrule(lr){2-5} \cmidrule(lr){6-9}
        & \multicolumn{2}{c}{Full} & \multicolumn{2}{c}{Adaptive} & \multicolumn{2}{c}{Full} & \multicolumn{2}{c}{Adaptive}\\
        \cmidrule(lr){2-3} \cmidrule(lr){4-5} \cmidrule(lr){6-7} \cmidrule(lr){8-9} 
        Model & Hourly & CDD & Hourly & CDD & Hourly & CDD & Hourly & CDD \\
        \midrule \\[-1.3ex]
        XGBoost 
        & 0.7\(\pm\)0.06 & 0.69\(\pm\)0.06 & \textbf{0.67\(\pm\)0.05} & 0.7\(\pm\)0.06
        & 2.485\(\pm\)0.21 & 0.022\(\pm\)0.002 & 1.688\(\pm\)0.133 & \textbf{0.016\(\pm\)0.001} \\
        FC-MLP 
        & 0.95\(\pm\)0.09 & 0.84\(\pm\)0.08 & 0.82\(\pm\)0.07 & 0.85\(\pm\)0.08
        & 4.5\(\pm\)0.429 & 0.036\(\pm\)0.003 & 1.564\(\pm\)0.134 & 0.089\(\pm\)0.009 \\
        Transformer 
        & 0.94\(\pm\)0.09 & 0.82\(\pm\)0.07 & 0.8\(\pm\)0.07 & 0.77\(\pm\)0.07
        & 7.186\(\pm\)0.697 & 1.086\(\pm\)0.096 & 2.449\(\pm\)0.206 & 0.022\(\pm\)0.002 \\
        LSTM 
        & 0.94\(\pm\)0.1 & 0.9\(\pm\)0.09 & 0.81\(\pm\)0.07 & 0.73\(\pm\)0.06
        & 7.346\(\pm\)0.824 & 0.126\(\pm\)0.013 & 3.243\(\pm\)0.277 & 0.026\(\pm\)0.002\\
        GFS 
        & \multicolumn{4}{c}{\textbf{\textit{0.5\(\pm\)0.04}}}
        & \multicolumn{4}{c}{-} \\        
    \end{tabular}%
    }
    \adjustbox{max width=\textwidth}{%
    \begin{tabular}{@{}lcccc|cccc@{}}
        \toprule
        & & \multicolumn{6}{c}{Wind speed, km/h} \\
        \cmidrule(lr){2-9}
        & \multicolumn{4}{c}{RMSE} & \multicolumn{4}{c}{RMSE\(\cdot\)Wh/prediction} \\
        \cmidrule(lr){2-5} \cmidrule(lr){6-9}
        & \multicolumn{2}{c}{Full} & \multicolumn{2}{c}{Adaptive} & \multicolumn{2}{c}{Full} & \multicolumn{2}{c}{Adaptive}\\
        \cmidrule(lr){2-3} \cmidrule(lr){4-5} \cmidrule(lr){6-7} \cmidrule(lr){8-9} 
        Model & Hourly & CDD & Hourly & CDD & Hourly & CDD & Hourly & CDD \\
        \midrule \\[-1.3ex]
        XGBoost 
        & 2.87\(\pm\)0.12 & \textbf{2.82\(\pm\)0.1} & 2.98\(\pm\)0.13 & 3.02\(\pm\)0.15
        & 10.47\(\pm\)0.44 & 0.09\(\pm\)0.0 & 8.6\(\pm\)0.36 & 0.08\(\pm\)0.0 \\
        FC-MLP 
        & 3.99\(\pm\)0.29 & 3.99\(\pm\)0.29 & 3.57\(\pm\)0.22 & 3.63\(\pm\)0.24
        & 14.31\(\pm\)1.05 & 14.31\(\pm\)1.05 & 5.78\(\pm\)0.36 & \textbf{0.05\(\pm\)0.0} \\
        Transformer 
        & 3.72\(\pm\)0.24 & 3.46\(\pm\)0.2 & 3.84\(\pm\)0.27 & 3.86\(\pm\)0.31
        & 28.78\(\pm\)1.87 & 0.49\(\pm\)0.03 & 11.83\(\pm\)0.82 & 1.3\(\pm\)0.1 \\
        LSTM 
        & 4.27\(\pm\)0.41 & 4.27\(\pm\)0.41 & 4.17\(\pm\)0.35 & 3.63\(\pm\)0.22
        & 26.03\(\pm\)2.47 & 26.27\(\pm\)2.5 & 14.07\(\pm\)1.18 & 0.11\(\pm\)0.01 \\
        GFS 
        & \multicolumn{4}{c}{\textbf{\textit{2.53\(\pm\)0.09}}}
        & \multicolumn{4}{c}{-} \\        
    \end{tabular}%
    }
    \adjustbox{max width=\textwidth}{%
    \begin{tabular}{@{}lcccc|cccc@{}}
        \toprule
        & & \multicolumn{6}{c}{Cloud cover, \%} \\
        \cmidrule(lr){2-9}
        & \multicolumn{4}{c}{RMSE} & \multicolumn{4}{c}{RMSE\(\cdot\)Wh/prediction} \\
        \cmidrule(lr){2-5} \cmidrule(lr){6-9}
        & \multicolumn{2}{c}{Full} & \multicolumn{2}{c}{Adaptive} & \multicolumn{2}{c}{Full} & \multicolumn{2}{c}{Adaptive}\\
        \cmidrule(lr){2-3} \cmidrule(lr){4-5} \cmidrule(lr){6-7} \cmidrule(lr){8-9} 
        Model & Hourly & CDD & Hourly & CDD & Hourly & CDD & Hourly & CDD \\
        \midrule \\[-1.3ex]
        XGBoost 
        & \textbf{34.87\(\pm\)0.81} & 35.78\(\pm\)0.85 & 36.64\(\pm\)1.13 & 36.96\(\pm\)1.28
        & 127.79\(\pm\)2.96 & 3.58\(\pm\)0.08 & 112.27\(\pm\)3.45 & \textbf{3.09\(\pm\)0.11} \\
        FC-MLP 
        & 45.4\(\pm\)1.3 & 44.31\(\pm\)1.41 & 49.26\(\pm\)2.38 & 47.46\(\pm\)2.22
        & 215.6\(\pm\)6.15 & 17.21\(\pm\)0.55 & 85.92\(\pm\)4.15 & 12.8\(\pm\)0.6 \\
        Transformer 
        & 38.04\(\pm\)0.8 & 37.31\(\pm\)0.79 & 45.57\(\pm\)1.9 & 47.94\(\pm\)2.82
        & 297.19\(\pm\)6.28 & 21.2\(\pm\)0.45 & 139.51\(\pm\)5.82 & 13.34\(\pm\)0.78 \\
        LSTM 
        & 47.58\(\pm\)0.65 & 47.58\(\pm\)0.65 & 46.4\(\pm\)2.03 & 48.64\(\pm\)1.91
        & 348.34\(\pm\)4.76 & 351.51\(\pm\)4.8 & 167.15\(\pm\)7.31 & 22.3\(\pm\)0.88 \\
        GFS 
        & \multicolumn{4}{c}{42.39\(\pm\)1.28}
        & \multicolumn{4}{c}{-} \\        
    \end{tabular}%
    }
    \adjustbox{max width=\textwidth}{%
    \begin{tabular}{@{}lcccc|cccc@{}}
        \toprule
        \large \textbf{Boston} & & \multicolumn{6}{c}{Temperature, \degree C} \\
        \cmidrule(lr){2-9}
        & \multicolumn{4}{c}{RMSE} & \multicolumn{4}{c}{RMSE\(\cdot\)Wh/prediction} \\
        \cmidrule(lr){2-5} \cmidrule(lr){6-9}
        & \multicolumn{2}{c}{Full} & \multicolumn{2}{c}{Adaptive} & \multicolumn{2}{c}{Full} & \multicolumn{2}{c}{Adaptive}\\
        \cmidrule(lr){2-3} \cmidrule(lr){4-5} \cmidrule(lr){6-7} \cmidrule(lr){8-9} 
        Model & Hourly & CDD & Hourly & CDD & Hourly & CDD & Hourly & CDD \\
        \midrule \\[-1.3ex]
        XGBoost 
        & 1.77\(\pm\)0.27 & 1.76\(\pm\)0.28 & 1.67\(\pm\)0.23 & 1.71\(\pm\)0.25
        & 5.76\(\pm\)0.87 & 0.16\(\pm\)0.02 & 4.0\(\pm\)0.53 & \textbf{0.04\(\pm\)0.0} \\
        FC-MLP 
        & 1.7\(\pm\)0.23 & \textbf{1.63\(\pm\)0.22} & 1.93\(\pm\)0.26 & 1.78\(\pm\)0.24
        & 7.68\(\pm\)1.05 & 4.6\(\pm\)0.62 & 3.61\(\pm\)0.48 & 1.86\(\pm\)0.24 \\
        Transformer 
        & 1.68\(\pm\)0.24 & 1.73\(\pm\)0.25 & 1.87\(\pm\)0.25 & 1.9\(\pm\)0.27
        & 12.9\(\pm\)1.86 & 4.97\(\pm\)0.71 & 5.56\(\pm\)0.72 & 0.67\(\pm\)0.09 \\
        LSTM 
        & 1.82\(\pm\)0.27 & 1.64\(\pm\)0.24 & 1.95\(\pm\)0.28 & 1.78\(\pm\)0.24
        & 13.83\(\pm\)2.03 & 4.74\(\pm\)0.67 & 7.78\(\pm\)1.11 & 1.22\(\pm\)0.16 \\
        GFS 
        & \multicolumn{4}{c}{\textbf{\textit{0.68\(\pm\)0.08}}}
        & \multicolumn{4}{c}{-} \\        
    \end{tabular}%
    }
    \adjustbox{max width=\textwidth}{%
    \begin{tabular}{@{}lcccc|cccc@{}}
        \toprule
        & & \multicolumn{6}{c}{Wind speed, km/h} \\
        \cmidrule(lr){2-9}
        & \multicolumn{4}{c}{RMSE} & \multicolumn{4}{c}{RMSE\(\cdot\)Wh/prediction} \\
        \cmidrule(lr){2-5} \cmidrule(lr){6-9}
        & \multicolumn{2}{c}{Full} & \multicolumn{2}{c}{Adaptive} & \multicolumn{2}{c}{Full} & \multicolumn{2}{c}{Adaptive}\\
        \cmidrule(lr){2-3} \cmidrule(lr){4-5} \cmidrule(lr){6-7} \cmidrule(lr){8-9} 
        Model & Hourly & CDD & Hourly & CDD & Hourly & CDD & Hourly & CDD \\
        \midrule \\[-1.3ex]
        XGBoost 
        & 3.94\(\pm\)0.26 & \textbf{3.91\(\pm\)0.26} & 4.21\(\pm\)0.3 & 4.07\(\pm\)0.3
        & 13.88\(\pm\)0.88 & 0.25\(\pm\)0.02 & 12.11\(\pm\)0.84 & \textbf{0.11\(\pm\)0.01} \\
        FC-MLP 
        & 4.91\(\pm\)0.47 & 5.01\(\pm\)0.49 & 4.84\(\pm\)0.37 & 4.77\(\pm\)0.36
        & 20.5\(\pm\)1.94 & 1.33\(\pm\)0.13 & 9.01\(\pm\)0.68 & 2.74\(\pm\)0.2 \\
        Transformer 
        & 4.63\(\pm\)0.41 & 4.63\(\pm\)0.43 & 4.49\(\pm\)0.35 & 4.55\(\pm\)0.36
        & 35.14\(\pm\)3.02 & 0.64\(\pm\)0.06 & 13.92\(\pm\)1.06 & 0.26\(\pm\)0.02 \\
        LSTM 
        & 5.61\(\pm\)0.56 & 5.61\(\pm\)0.56 & 4.8\(\pm\)0.4 & 4.92\(\pm\)0.42
        & 31.36\(\pm\)3.1 & 31.36\(\pm\)3.1 & 15.97\(\pm\)1.33 & 11.76\(\pm\)1.0 \\
        GFS 
        & \multicolumn{4}{c}{\textbf{\textit{2.78\(\pm\)0.12}}}
        & \multicolumn{4}{c}{-} \\        
    \end{tabular}%
    }
    \adjustbox{max width=\textwidth}{%
    \begin{tabular}{@{}lcccc|cccc@{}}
        \toprule
        & & \multicolumn{6}{c}{Cloud cover, \%} \\
        \cmidrule(lr){2-9}
        & \multicolumn{4}{c}{RMSE} & \multicolumn{4}{c}{RMSE\(\cdot\)Wh/prediction} \\
        \cmidrule(lr){2-5} \cmidrule(lr){6-9}
        & \multicolumn{2}{c}{Full} & \multicolumn{2}{c}{Adaptive} & \multicolumn{2}{c}{Full} & \multicolumn{2}{c}{Adaptive}\\
        \cmidrule(lr){2-3} \cmidrule(lr){4-5} \cmidrule(lr){6-7} \cmidrule(lr){8-9} 
        Model & Hourly & CDD & Hourly & CDD & Hourly & CDD & Hourly & CDD \\
        \midrule \\[-1.3ex]
        XGBoost 
        & 37.0\(\pm\)1.54 & 36.51\(\pm\)1.61 & 37.03\(\pm\)1.54 & \textbf{36.28\(\pm\)1.53}
        & 128.55\(\pm\)5.35 & 2.31\(\pm\)0.1 & 111.37\(\pm\)4.55 & \textbf{1.99\(\pm\)0.08} \\
        FC-MLP 
        & 40.78\(\pm\)1.56 & 40.15\(\pm\)1.57 & 46.29\(\pm\)2.55 & 46.09\(\pm\)2.49
        & 170.12\(\pm\)6.54 & 15.25\(\pm\)0.6 & 69.16\(\pm\)3.76 & 26.9\(\pm\)1.44 \\
        Transformer 
        & 37.99\(\pm\)1.67 & 37.92\(\pm\)1.66 & 41.52\(\pm\)1.81 & 38.88\(\pm\)1.33
        & 286.2\(\pm\)12.13 & 270.14\(\pm\)11.34 & 130.87\(\pm\)5.64 & 12.29\(\pm\)0.42 \\
        LSTM 
        & 45.61\(\pm\)1.75 & 45.61\(\pm\)1.75 & 49.43\(\pm\)2.82 & 49.43\(\pm\)2.82
        & 299.71\(\pm\)11.58 & 299.71\(\pm\)11.58 & 143.6\(\pm\)8.03 & 143.6\(\pm\)8.03 \\
        GFS 
        & \multicolumn{4}{c}{\textbf{\textit{28.95\(\pm\)1.35}}}
        & \multicolumn{4}{c}{-} \\        
    \end{tabular}%
    }
\end{table}

\begin{figure}[h]
    \centering
    \includegraphics[width=1\linewidth]{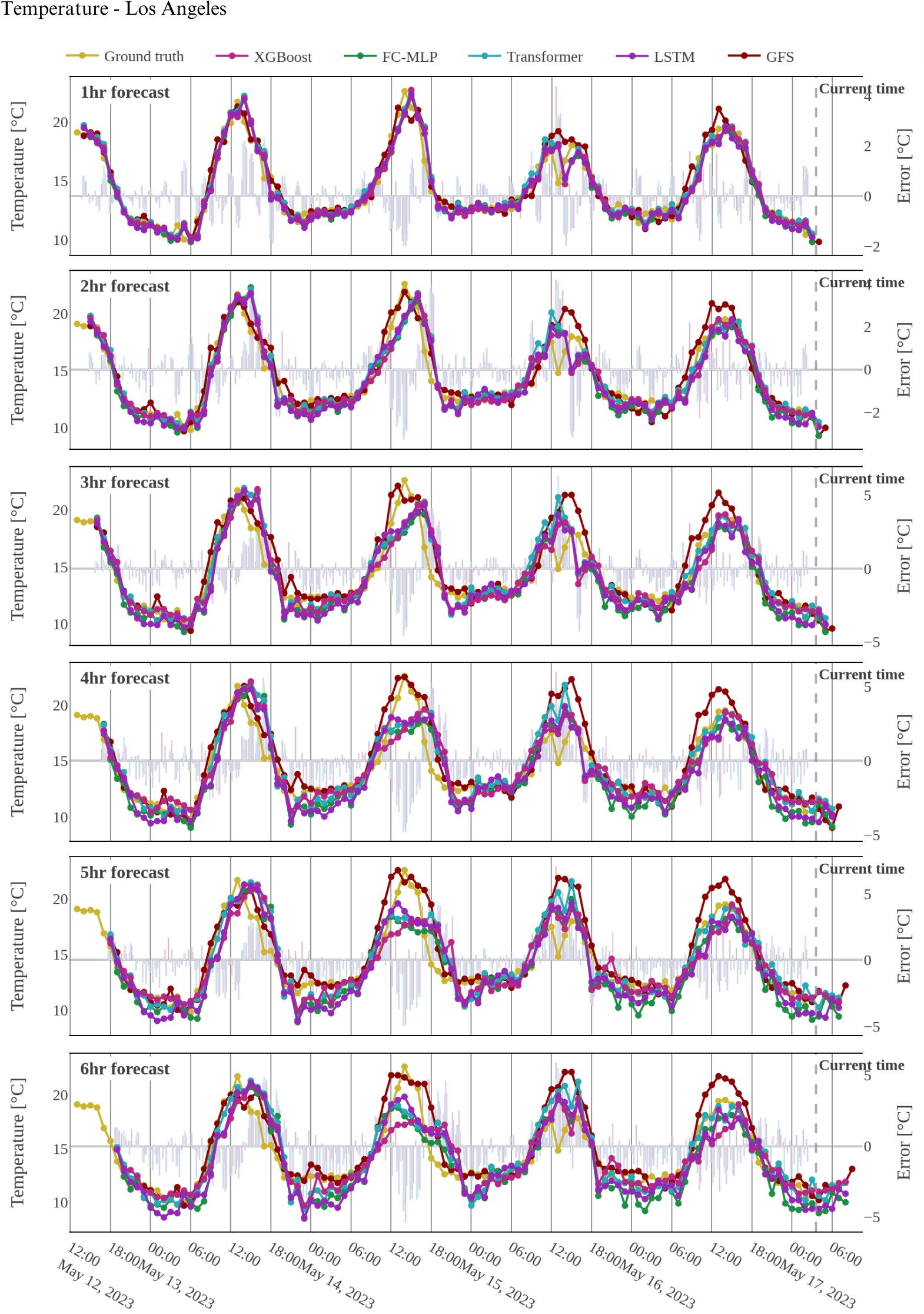}
    \caption{Temperature forecasts for Los Angeles using hourly retraining on the full training data. Rows show forecasts for separate forecast horizons (1hr-6hr). Errors (difference between forecast and \emph{Ground truth}) are indicated on the right-hand y-axis and shown as semi-transparent bars.
    \label{sup-fig:temp-la}}
\end{figure}

\begin{figure}[h]
    \centering
    \includegraphics[width=1\linewidth]{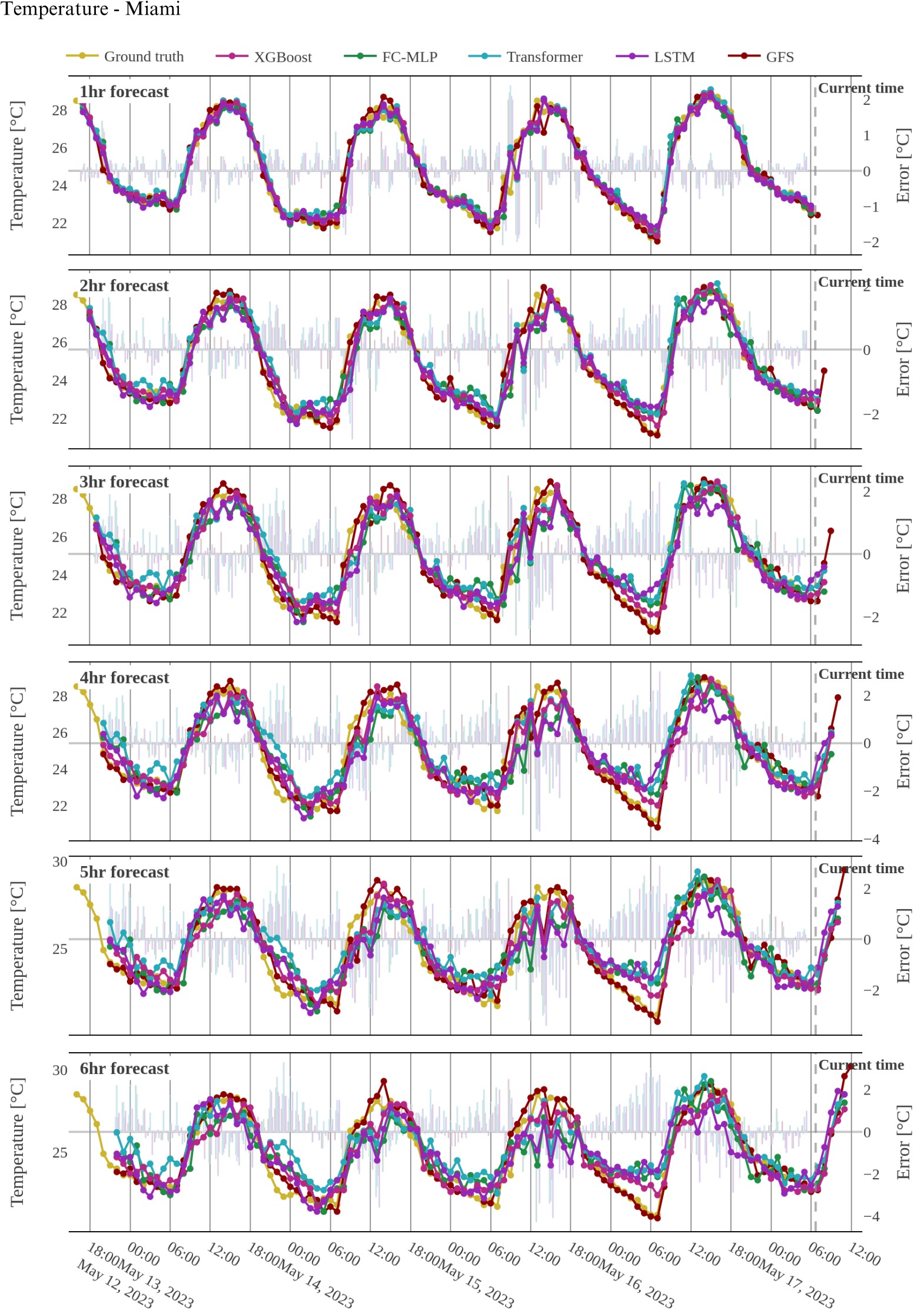}
    \caption{Temperature forecasts for Miami using hourly retraining on the full training data. Rows show forecasts for separate forecast horizons (1hr-6hr). Errors (difference between forecast and \emph{Ground truth}) are indicated on the right-hand y-axis and shown as semi-transparent bars.
    \label{sup-fig:temp-mia}}
\end{figure}

\begin{figure}[h]
    \centering
    \includegraphics[width=1\linewidth]{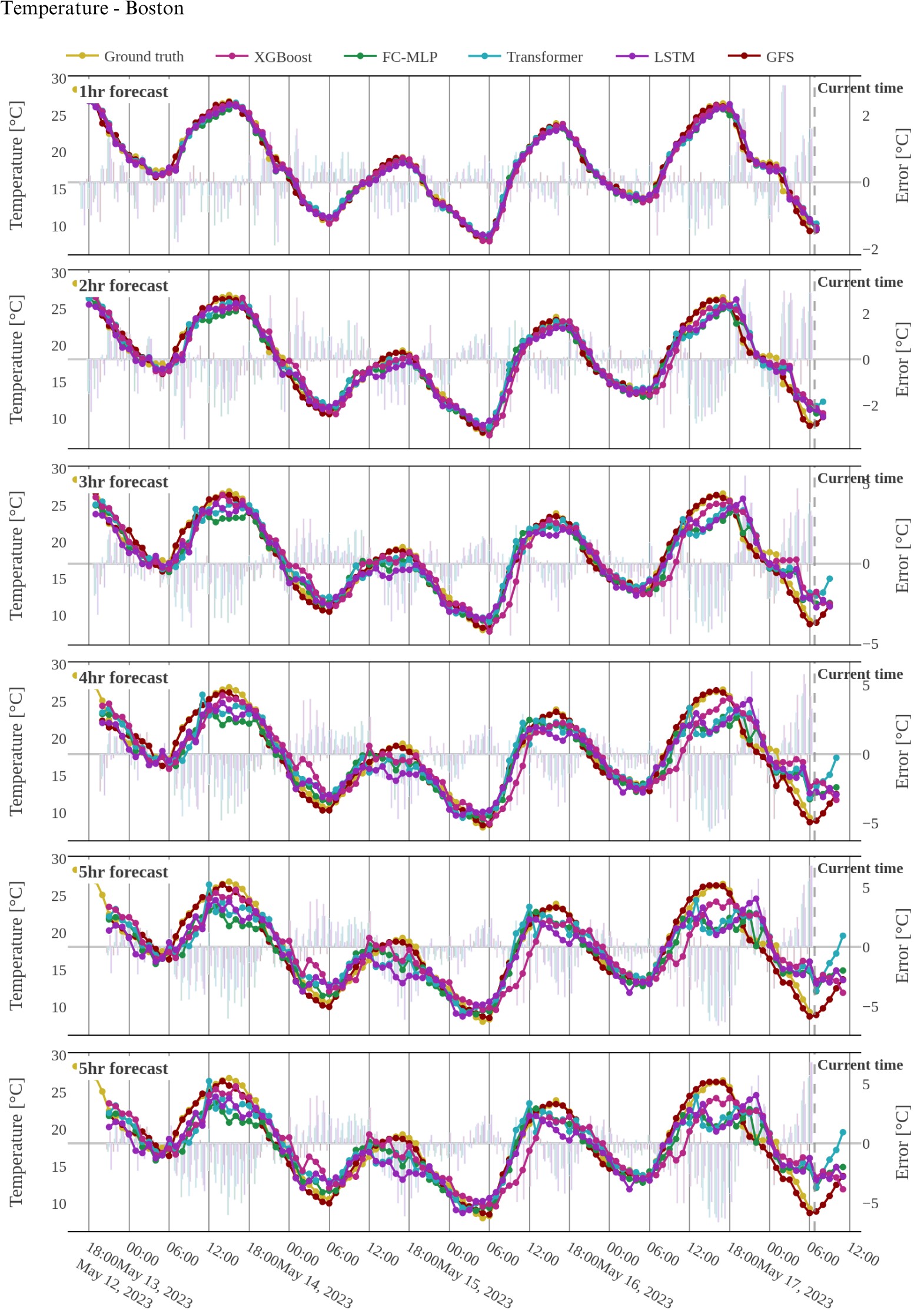}
    \caption{Temperature forecasts for Boston using hourly retraining on the full training data. Rows show forecasts for separate forecast horizons (1hr-6hr). Errors (difference between forecast and \emph{Ground truth}) are indicated on the right-hand y-axis and shown as semi-transparent bars.
    \label{sup-fig:temp-bos}}
\end{figure}

\begin{figure}[h]
    \centering
    \includegraphics[width=1\linewidth]{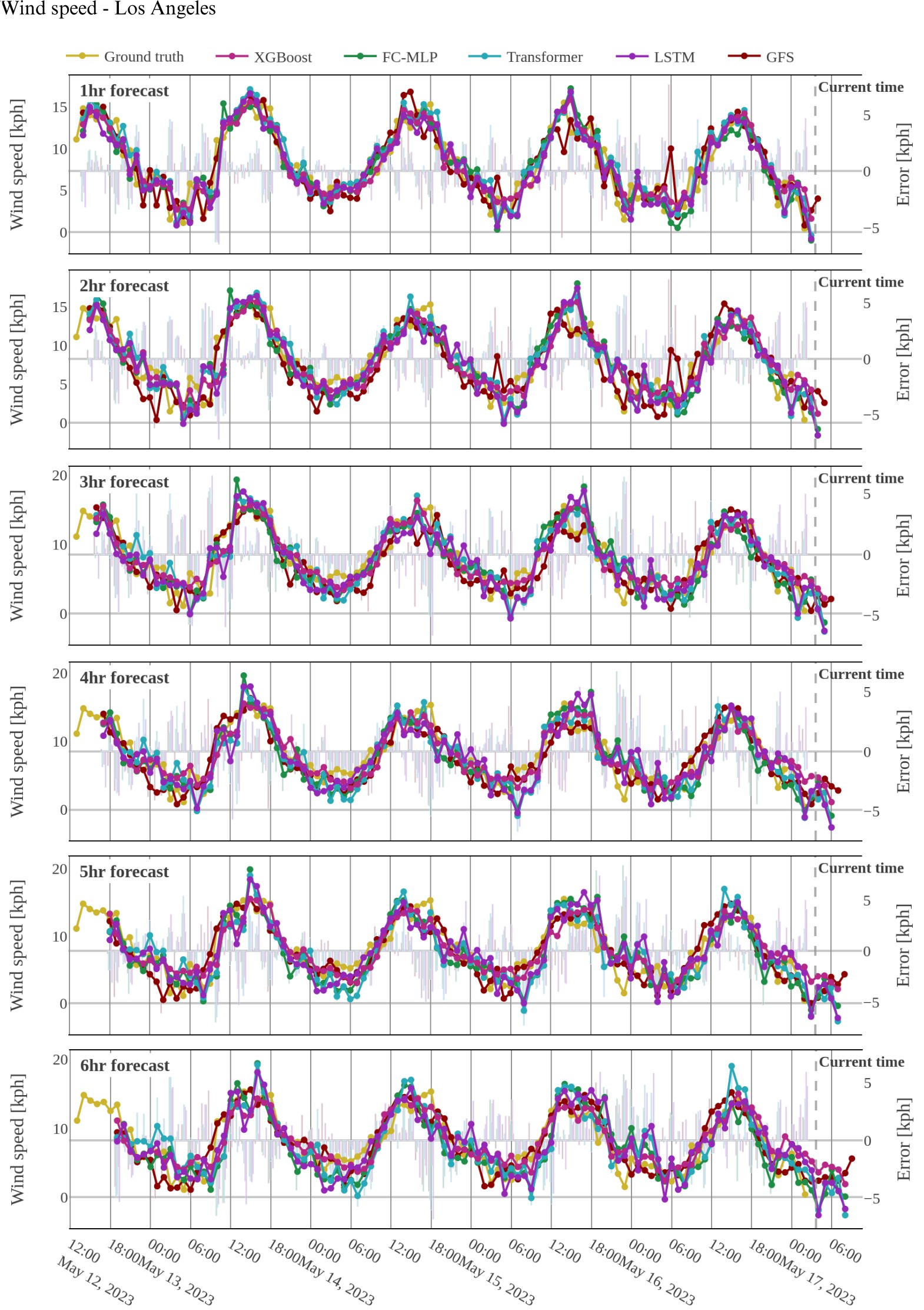}
    \caption{Wind speed forecasts for Los Angeles using hourly retraining on the full training data. Rows show forecasts for separate forecast horizons (1hr-6hr). Errors (difference between forecast and \emph{Ground truth}) are indicated on the right-hand y-axis and shown as semi-transparent bars.
    \label{sup-fig:ws-la}}
\end{figure}

\begin{figure}[h]
    \centering
    \includegraphics[width=1\linewidth]{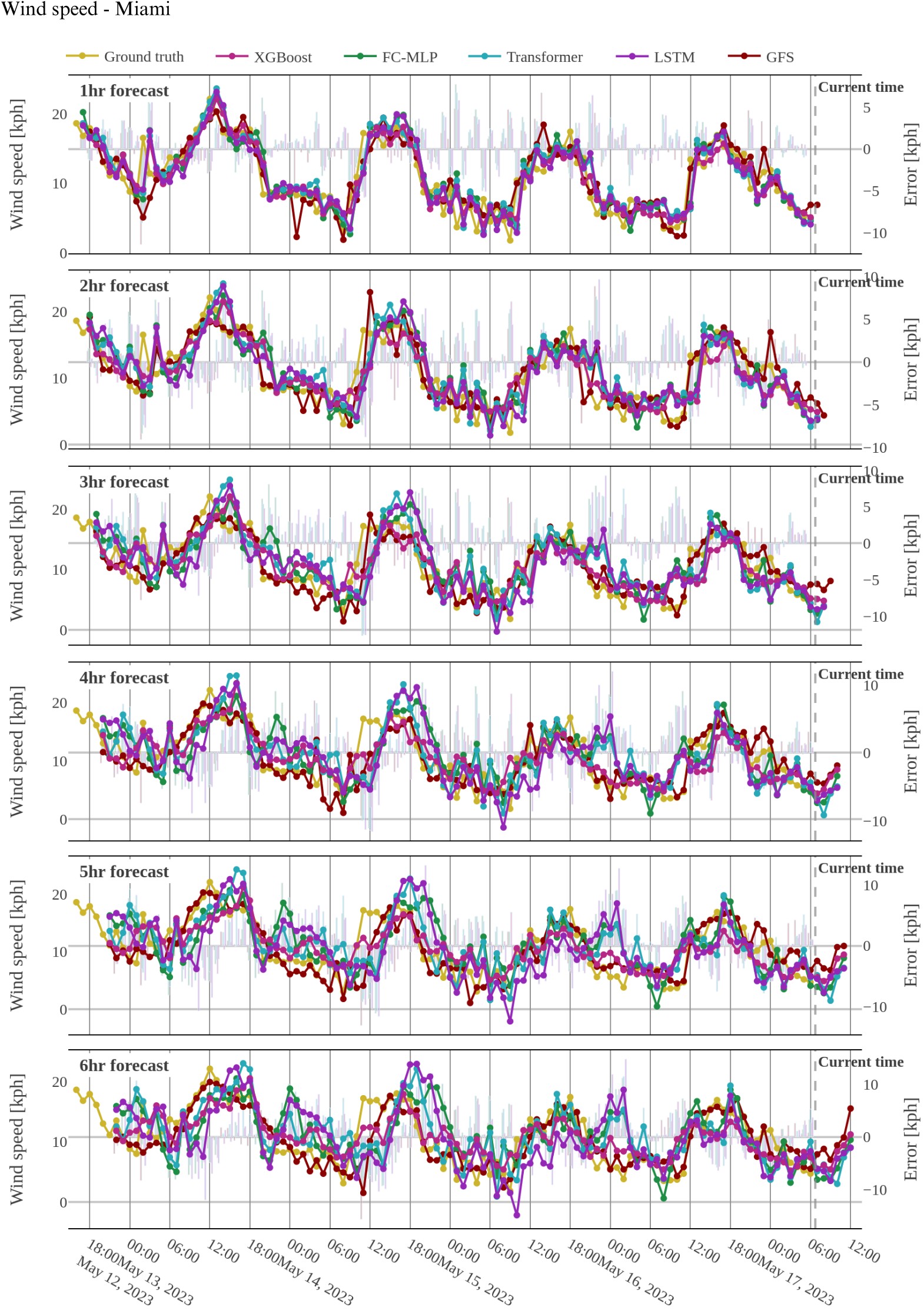}
    \caption{Wind speed forecasts for Miami using hourly retraining on the full training data. Rows show forecasts for separate forecast horizons (1hr-6hr). Errors (difference between forecast and \emph{Ground truth}) are indicated on the right-hand y-axis and shown as semi-transparent bars.
    \label{sup-fig:ws-mia}}
\end{figure}

\begin{figure}[h]
    \centering
    \includegraphics[width=1\linewidth]{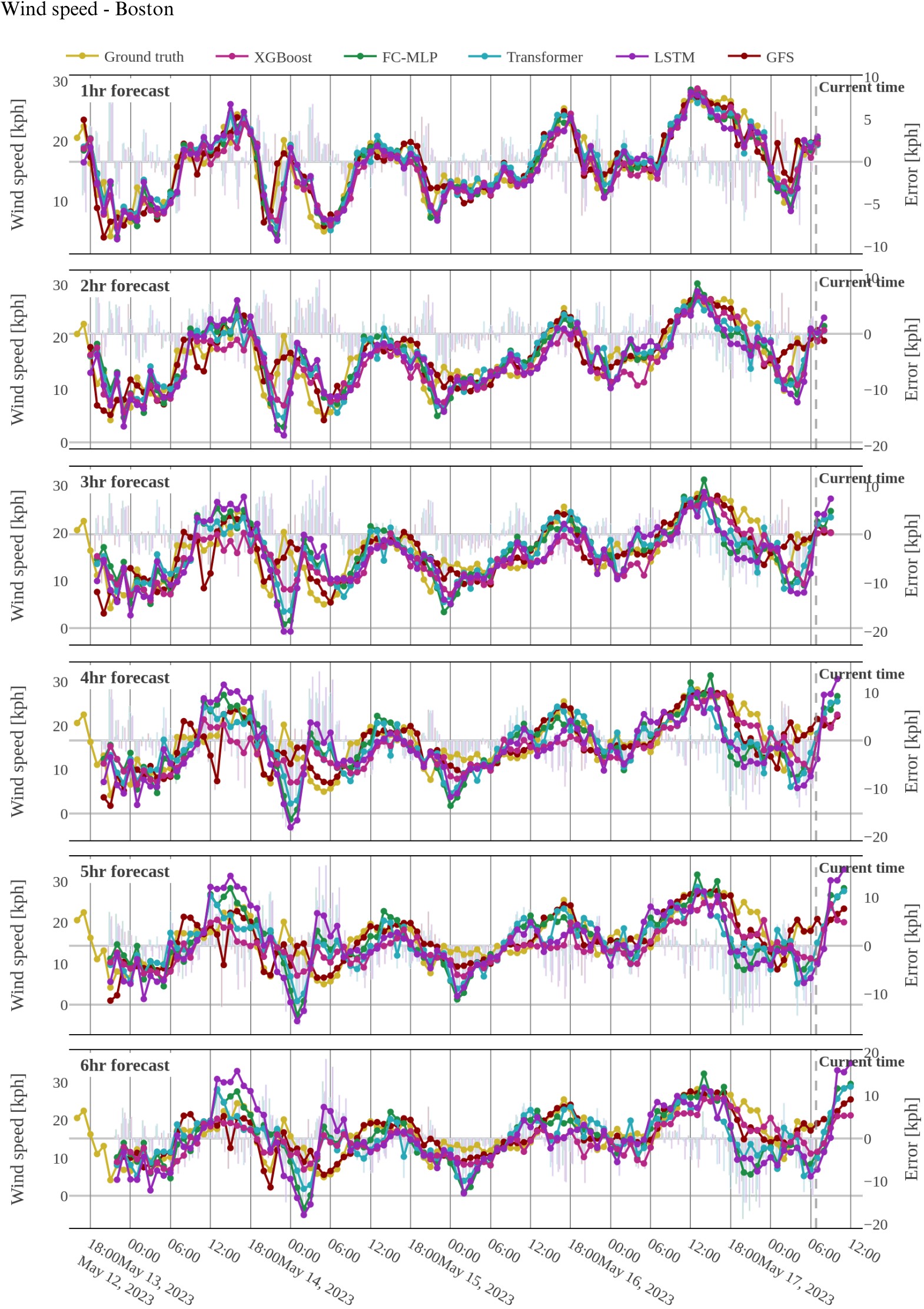}
    \caption{Wind speed forecasts for Boston using hourly retraining on the full training data. Rows show forecasts for separate forecast horizons (1hr-6hr). Errors (difference between forecast and \emph{Ground truth}) are indicated on the right-hand y-axis and shown as semi-transparent bars.
    \label{sup-fig:ws-bos}}
\end{figure}

\begin{figure}[h]
    \centering
    \includegraphics[width=1\linewidth]{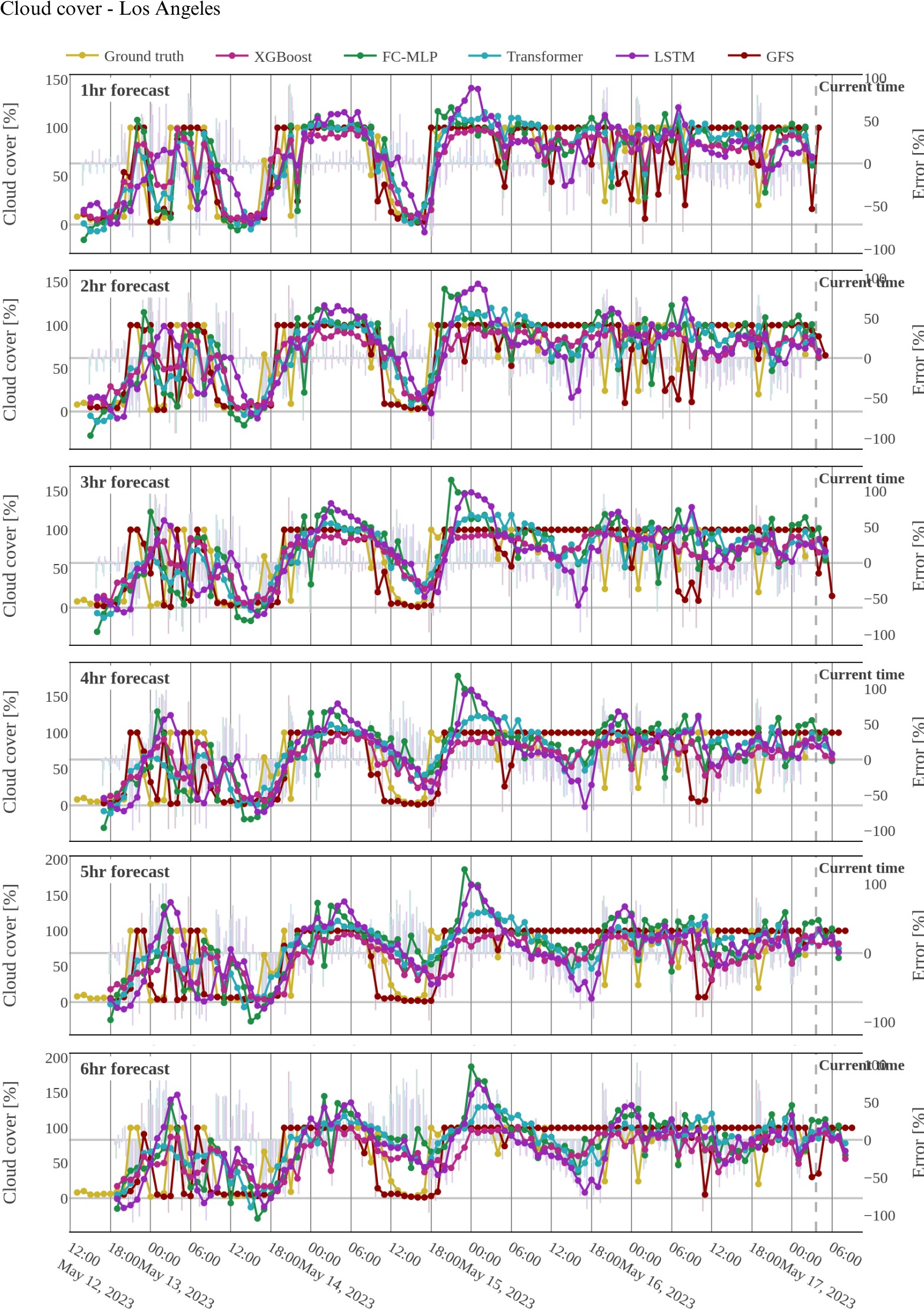}
    \caption{Cloud cover forecasts for Los Angeles using hourly retraining on the full training data. Rows show forecasts for separate forecast horizons (1hr-6hr). Errors (difference between forecast and \emph{Ground truth}) are indicated on the right-hand y-axis and shown as semi-transparent bars.
    \label{sup-fig:cc-la}}
\end{figure}

\begin{figure}[h]
    \centering
    \includegraphics[width=1\linewidth]{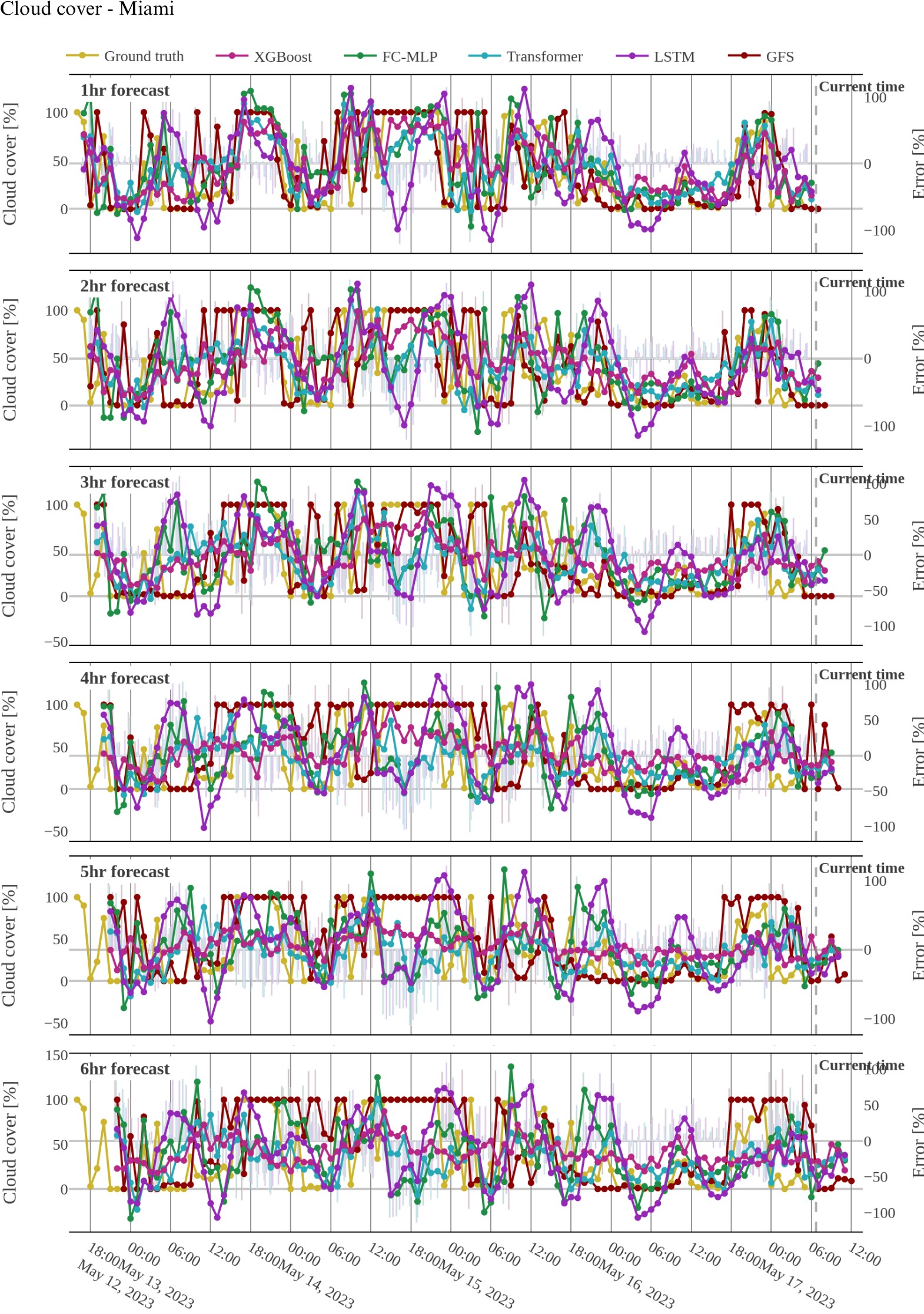}
    \caption{Cloud cover forecasts for Miami using hourly retraining on the full training data. Rows show forecasts for separate forecast horizons (1hr-6hr). Errors (difference between forecast and \emph{Ground truth}) are indicated on the right-hand y-axis and shown as semi-transparent bars.
    \label{sup-fig:cc-mia}}
\end{figure}

\begin{figure}[h]
    \centering
    \includegraphics[width=1\linewidth]{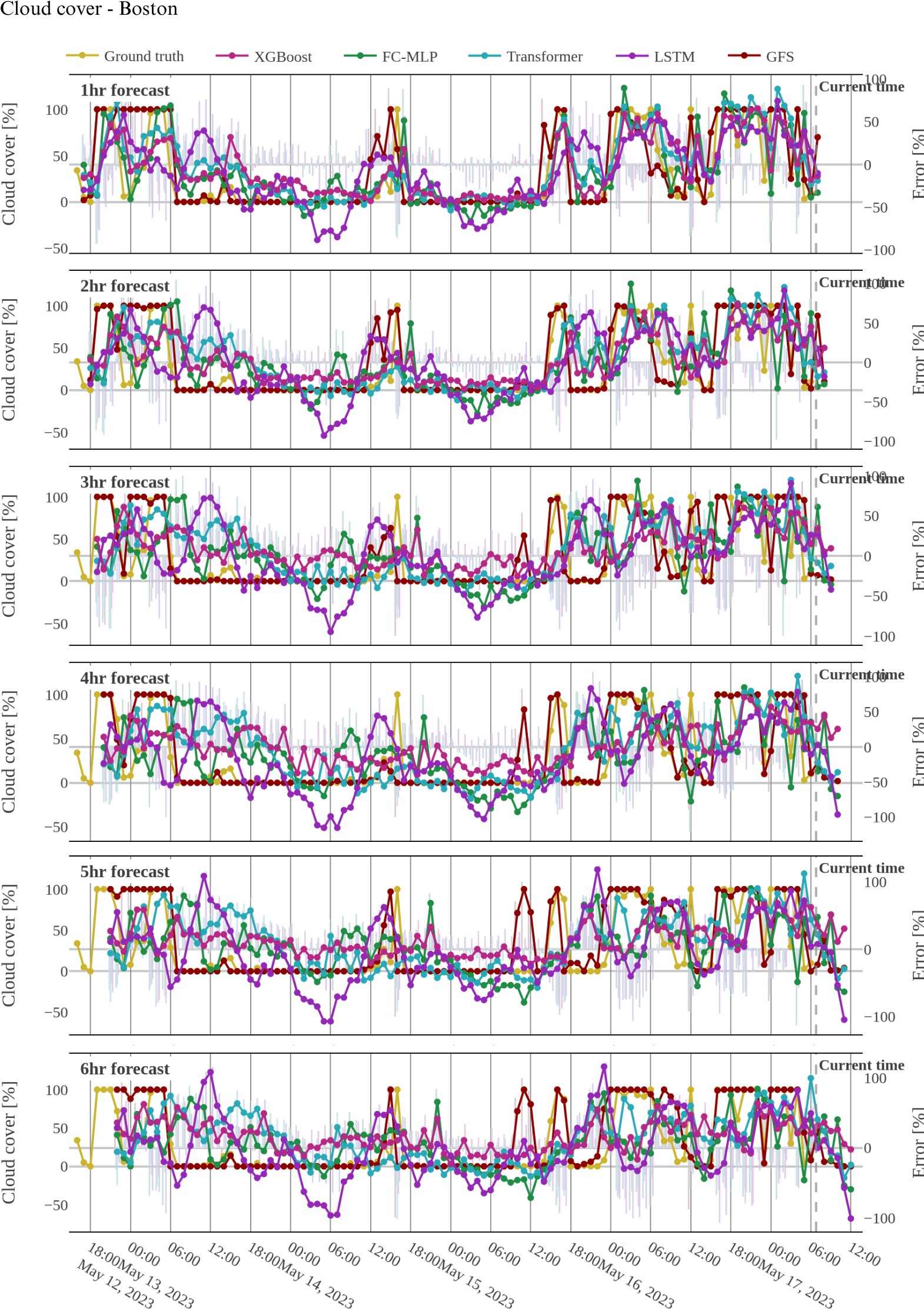}
    \caption{Cloud cover forecasts for Boston using hourly retraining on the full training data. Rows show forecasts for separate forecast horizons (1hr-6hr). Errors (difference between forecast and \emph{Ground truth}) are indicated on the right-hand y-axis and shown as semi-transparent bars.
    \label{sup-fig:cc-bos}}
\end{figure}

\begin{table}[h!]
    \centering
    \caption{Minimum validation loss and expended power for temperature forecasts using the full and adaptive training window.
    \label{sup-tab:val-pwr-temp}}
    \adjustbox{max width=\textwidth}{%
    \begin{tabular}{@{}lccccc@{}}
        \toprule
        & & \multicolumn{4}{c}{Temperature, \degree C} \\
        \cmidrule(lr){3-6}
        & & \multicolumn{2}{c}{Full} & \multicolumn{2}{c}{Adaptive} \\
        \cmidrule(lr){3-4} \cmidrule(lr){5-6}
        & Model & Pwr expended & Min. val. loss & Pwr expended & Min. val. loss\\
        \midrule \\[-1.3ex]
        \multirow{4}{*}{\rotatebox[origin=c]{90}{Los Angeles}}
        & XGBoost 
         & 3.35 $\pm$ 0.02 & 0.0031 $\pm$ 5.9e-04  
         & 2.23 $\pm$ 0.33 & 0.0042 $\pm$ 5.9e-04 \\[.3ex]
        & FC-MLP  
         & 4.79 $\pm$ 0.37 & \textbf{0.0029 $\pm$ 6.4e-04} 
         & \textbf{1.75 $\pm$ 0.5} & 0.0042 $\pm$ 6.6e-04\\[.3ex]
        & Transformer  
         & 7.73 $\pm$ 0.19 & 0.0029 $\pm$ 7.2e-04 
         & 2.79 $\pm$ 0.76 & 0.004 $\pm$ 7.9e-04\\[.3ex]
        & LSTM 
         & 8.03 $\pm$ 0.73 & 0.0031 $\pm$ 5.7e-04 
         & 3.6 $\pm$ 0.74 & 0.0044 $\pm$ 1.0e-03 \\[1.3ex]
        \midrule \\[-1.3ex]
        \multirow{4}{*}{\rotatebox[origin=c]{90}{Miami}}
        & XGBoost 
         & 3.55 $\pm$ 0.02 & \textbf{0.0018 $\pm$ 8.5e-05}  
         & 2.5 $\pm$ 0.3 & 0.0026 $\pm$ 6.1e-04 \\[.3ex]
        & FC-MLP 
         & 4.73 $\pm$ 0.35 & 0.0023 $\pm$ 1.1e-04 
         & \textbf{1.9 $\pm$ 0.42} & 0.0031 $\pm$ 2.7e-04  \\[.3ex]
        & Transformer 
         & 7.66 $\pm$ 0.31 & 0.0022 $\pm$ 1.1e-04 
         & 3.06 $\pm$ 0.69 & 0.0028 $\pm$ 3.8e-04 \\[.3ex]
        & LSTM 
         & 7.77 $\pm$ 0.9 & 0.0028 $\pm$ 1.3e-04 
         & 3.98 $\pm$ 0.57 & 0.0034 $\pm$ 3.1e-04 \\[1.0ex]
        \midrule \\[-1.3ex]
        \multirow{4}{*}{\rotatebox[origin=c]{90}{Boston}}
        & XGBoost 
         & 3.33 $\pm$ 0.03 & 0.0057 $\pm$ 2.4e-04
         & 2.49 $\pm$ 0.14 & 0.0064 $\pm$ 5.1e-04   \\[.3ex]
        & FC-MLP 
         & 4.57 $\pm$ 0.49 & \textbf{0.0041 $\pm$ 2.0e-04} 
         & \textbf{1.9 $\pm$ 0.28} & 0.0057 $\pm$ 7.0e-04 \\[.3ex]
        & Transformer 
         & 7.85 $\pm$ 0.33 & 0.0041 $\pm$ 3.5e-04 
         & 3.09 $\pm$ 0.34 & 0.0054 $\pm$ 7.4e-04 \\[.3ex]
        & LSTM 
         & 7.73 $\pm$ 0.82 & 0.0047 $\pm$ 3.8e-04  
         & 4.08 $\pm$ 0.36 & 0.0062 $\pm$ 7.3e-04 \\[1.0ex]
    \end{tabular}%
    }
\end{table}

\begin{table}[h!]
    \centering
    \caption{Minimum validation loss and expended power for wind speed forecasts using the full and adaptive training window.
    \label{sup-tab:val-pwr-ws}}
    \adjustbox{max width=\textwidth}{%
    \begin{tabular}{@{}lccccc@{}}
        \toprule
        & & \multicolumn{4}{c}{Wind speed, km/h} \\
        \cmidrule(lr){3-6}
        & & \multicolumn{2}{c}{Full} & \multicolumn{2}{c}{Adaptive} \\
        \cmidrule(lr){3-4} \cmidrule(lr){5-6}
        & Model & Pwr expended & Min. val. loss & Pwr expended & Min. val. loss\\
        \midrule \\[-1.3ex]
        \multirow{4}{*}{\rotatebox[origin=c]{90}{Los Angeles}}
        & XGBoost  
         & 3.37 $\pm$ 0.03 & \textbf{0.0031 $\pm$ 4.5e-04} 
         & 2.52 $\pm$ 0.33 & 0.0061 $\pm$ 1.7e-03\\[.3ex]
        & FC-MLP  
         & 3.51 $\pm$ 0.49 & 0.0054 $\pm$ 6.5e-04 
         & \textbf{1.41 $\pm$ 0.39} & 0.0088 $\pm$ 2.1e-03 \\[.3ex]
        & Transformer  
         & 6.8 $\pm$ 0.56 & 0.0048 $\pm$ 4.9e-04 
         & 2.7 $\pm$ 0.66 & 0.0083 $\pm$ 1.9e-03 \\[.3ex]
        & LSTM 
         & 5.4 $\pm$ 0.39 & 0.0051 $\pm$ 5.6e-04 
         & 2.92 $\pm$ 0.44 & 0.0087 $\pm$ 2.8e-03 \\[1.3ex]
        \midrule \\[-1.3ex]
        \multirow{4}{*}{\rotatebox[origin=c]{90}{Miami}}
        & XGBoost 
         & 3.65 $\pm$ 0.02 & \textbf{0.0061 $\pm$ 2.6e-04}
         & 2.88 $\pm$ 0.24 & 0.0079 $\pm$ 8.4e-04  \\[.3ex]
        & FC-MLP 
         & 3.59 $\pm$ 0.61 & 0.0073 $\pm$ 2.8e-04  
         & \textbf{1.62 $\pm$ 0.36} & 0.0086 $\pm$ 6.3e-04 \\[.3ex]
        & Transformer  
         & 7.75 $\pm$ 0.24 & 0.007 $\pm$ 3.2e-04 
         & 3.08 $\pm$ 0.71 & 0.0083 $\pm$ 8.0e-04 \\[.3ex]
        & LSTM 
         & 6.09 $\pm$ 0.78 & 0.0076 $\pm$ 3.2e-04
         & 3.37 $\pm$ 0.5 & 0.0088 $\pm$ 6.8e-04  \\[1.0ex]
        \midrule \\[-1.3ex]
        \multirow{4}{*}{\rotatebox[origin=c]{90}{Boston}}
        & XGBoost 
         & 3.53 $\pm$ 0.03 & \textbf{0.0087 $\pm$ 1.9e-04} 
         & 2.89 $\pm$ 0.1 & 0.0112 $\pm$ 4.1e-04 \\[.3ex]
        & FC-MLP 
         & 4.19 $\pm$ 0.57 & 0.0112 $\pm$ 3.6e-04 
         & \textbf{1.86 $\pm$ 0.26} & 0.0132 $\pm$ 6.9e-04 \\[.3ex]
        & Transformer 
         & 7.61 $\pm$ 0.41 & 0.011 $\pm$ 4.4e-04 
         & 3.11 $\pm$ 0.29 & 0.0131 $\pm$ 6.4e-04 \\[.3ex]
        & LSTM 
         & 5.62 $\pm$ 0.55 & 0.0119 $\pm$ 4.6e-04 
         & 3.33 $\pm$ 0.37 & 0.0138 $\pm$ 6.1e-04 \\[1.0ex]
    \end{tabular}%
    }
\end{table}

\begin{table}[h!]
    \centering
    \caption{Minimum validation loss and expended power for cloud cover forecasts using the full and adaptive training window.
    \label{sup-tab:val-pwr-cc}}
    \adjustbox{max width=\textwidth}{%
    \begin{tabular}{@{}lccccc@{}}
        \toprule
        & & \multicolumn{4}{c}{Cloud cover, \%} \\
        \cmidrule(lr){3-6}
        & & \multicolumn{2}{c}{Full} & \multicolumn{2}{c}{Adaptive} \\
        \cmidrule(lr){3-4} \cmidrule(lr){5-6}
        & Model & Pwr expended & Min. val. loss & Pwr expended & Min. val. loss\\
        \midrule \\[-1.3ex]
        \multirow{4}{*}{\rotatebox[origin=c]{90}{Los Angeles}}
        & XGBoost  
         & 3.53 $\pm$ 0.02 & 0.0286 $\pm$ 7.5e-04 
         & 2.76 $\pm$ 0.39 & 0.0312 $\pm$ 1.9e-03 \\[.3ex]
        & FC-MLP  
         & 4.63 $\pm$ 0.42 & 0.0371 $\pm$ 1.2e-03 
         & \textbf{1.63 $\pm$ 0.46} & 0.0421 $\pm$ 1.6e-03 \\[.3ex]
        & Transformer  
         & 7.91 $\pm$ 0.22 & \textbf{0.0273 $\pm$ 9.3e-04} 
         & 2.79 $\pm$ 0.7 & 0.0391 $\pm$ 4.6e-03 \\[.3ex]
        & LSTM 
         & 8.02 $\pm$ 0.48 & 0.0342 $\pm$ 1.4e-03 
         & 3.03 $\pm$ 0.64 & 0.0431 $\pm$ 1.5e-03 \\[1.3ex]
        \midrule \\[-1.3ex]
        \multirow{4}{*}{\rotatebox[origin=c]{90}{Miami}}
        & XGBoost 
         & 3.66 $\pm$ 0.03 & \textbf{0.0256 $\pm$ 1.3e-03} 
         & 3.06 $\pm$ 0.23 & 0.0276 $\pm$ 1.8e-03 \\[.3ex]
        & FC-MLP 
         & 4.75 $\pm$ 0.35 & 0.0363 $\pm$ 4.8e-03 
         & \textbf{1.74 $\pm$ 0.55} & 0.0403 $\pm$ 5.6e-03 \\[.3ex]
        & Transformer  
         & 7.81 $\pm$ 0.23 & 0.0257 $\pm$ 3.1e-03 
         & 3.06 $\pm$ 0.71 & 0.0347 $\pm$ 7.4e-03 \\[.3ex]
        & LSTM 
         & 7.32 $\pm$ 0.42 & 0.0284 $\pm$ 3.3e-03 
         & 3.6 $\pm$ 0.86 & 0.0401 $\pm$ 5.4e-03 \\[1.0ex]
        \midrule \\[-1.3ex]
        \multirow{4}{*}{\rotatebox[origin=c]{90}{Boston}}
        & XGBoost 
         & 3.49 $\pm$ 0.03 & 0.0231 $\pm$ 5.4e-04 
         & 3.02 $\pm$ 0.12 & 0.0247 $\pm$ 1.1e-03 \\[.3ex]
        & FC-MLP 
         & 4.18 $\pm$ 0.67 & 0.0286 $\pm$ 1.9e-03 
         & \textbf{1.5 $\pm$ 0.35} & 0.0312 $\pm$ 2.8e-03 \\[.3ex]
        & Transformer 
         & 7.56 $\pm$ 0.32 & \textbf{0.0217 $\pm$ 1.8e-03} 
         & 3.16 $\pm$ 0.29 & 0.0252 $\pm$ 2.3e-03 \\[.3ex]
        & LSTM 
         & 6.57 $\pm$ 0.67 & 0.0273 $\pm$ 2.2e-03 
         & 2.93 $\pm$ 0.42 & 0.0308 $\pm$ 3.1e-03 \\[1.0ex]
    \end{tabular}%
    }
\end{table}

\end{document}